\documentclass{article}

\usepackage{microtype}
\usepackage{graphicx}
\usepackage{booktabs} % for professional tables

\usepackage{hyperref}

\usepackage[preprint]{icml2026}

\usepackage{amsmath}
\usepackage{amssymb}
\usepackage{mathtools}
\usepackage{amsthm}

\usepackage{svg}
\usepackage{subfig}

\usepackage[capitalize,noabbrev]{cleveref}

\theoremstyle{plain}

\theoremstyle{definition}

\theoremstyle{remark}

\usepackage[textsize=tiny]{todonotes}

\icmltitlerunning{Hierarchical Lead Critic based Multi-Agent Reinforcement Learning}

\begin{document}

\twocolumn[
  \icmltitle{Hierarchical Lead Critic based Multi-Agent Reinforcement Learning}



  \icmlsetsymbol{equal}{*}

  \begin{icmlauthorlist}
    \icmlauthor{David Eckel}{equal,comp}
    \icmlauthor{Henri Meeß}{comp}
  \end{icmlauthorlist}

  \icmlaffiliation{comp}{Fraunhofer Institute IVI, Ingolstadt, Germany}

  \icmlcorrespondingauthor{David Eckel}{david.eckel@ivi.fraunhofer.de}

  \icmlkeywords{Multi-Agent Reinforcement Learning, Hierarchical Reinforcement Learning, Partial Observability}

  \vskip 0.3in
]



\printAffiliationsAndNotice{}  % no special notice (required even if empty)

\begin{abstract}
Cooperative Multi-Agent Reinforcement Learning (MARL) solves complex tasks that require coordination from multiple agents, but is often limited to either local (independent learning) or global (centralized learning) perspectives. In this paper, we introduce a novel sequential training scheme and MARL architecture, which learns from multiple perspectives on different hierarchy levels. We propose the Hierarchical Lead Critic (HLC) - inspired by natural emerging distributions in team structures, where following high-level objectives combines with low-level execution. HLC demonstrates that introducing multiple hierarchies, leveraging local and global perspectives, can lead to improved performance with high sample efficiency and robust policies. Experimental results conducted on cooperative, non-communicative, and partially observable MARL benchmarks demonstrate that HLC outperforms single hierarchy baselines and scales robustly with increasing amounts of agents and difficulty.
\end{abstract}

\section{Introduction}

\begin{figure}[ht]
  \vskip 0.2in
  \begin{center}
    \centerline{\includegraphics[width=\columnwidth]{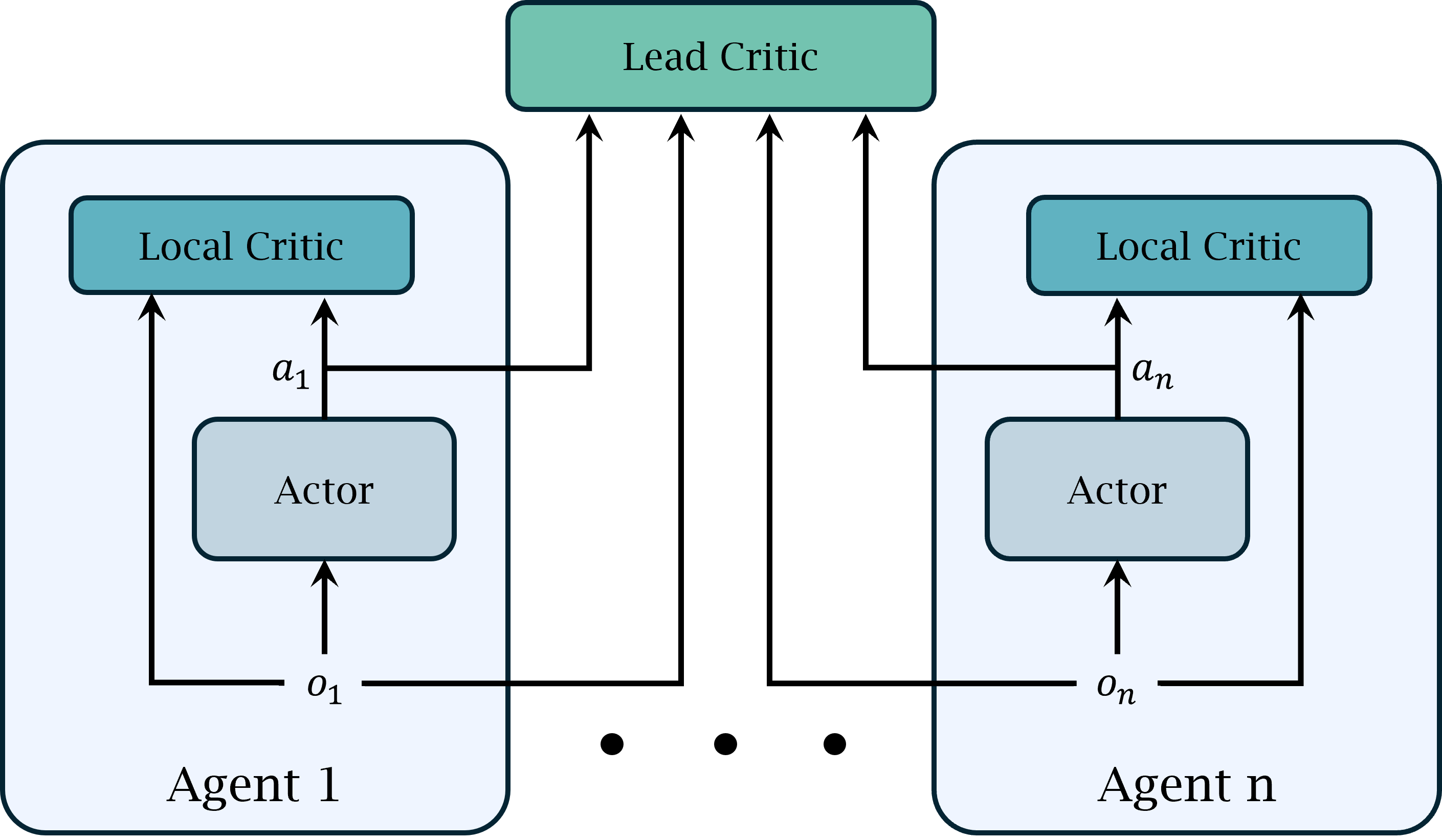}}
    \caption{
      Hierarchical Lead Critic (HLC) structure. Agents are evaluated by all related critics (local critics and Lead Critics) sequentially.
    }
    \label{fig:HLC-Structure}
  \end{center}
\end{figure}
Cooperative Multi-Agent Reinforcement Learning (MARL) addresses tasks that require coordinated behavior among multiple agents operating in shared environments, often paired with with partially observability. While recent state-of-the-art (SOTA) methods~\cite{MAT, HASAC, SABLE} leverage centralized information during training following the Centralized Training with Decentralized Execution (CTDE) paradigm~\citep{CTDE}, they typically optimize from a single perspective—either purely local (independent learning) or purely global (centralized learning). This overlooks, that for achieving a high-level objective often the individual agents need to learn a reasonable low-level policy first before optimizing execution. Following the idea that a team where individuals struggle to perform basic tasks will likely fail to effectively solve the given overall.\\
We introduce the Hierarchical Lead Critic (HLC), a MARL architecture paired with a sequential training scheme that learns from multiple perspectives across hierarchy levels. HLC organizes agents into groups, each guided by a Lead Critic that aggregates local observation–action information and provides group-level value estimates, while also retaining local critics at the agent level; note that a fully centralized critic is a special case of a Lead Critic with maximal receptive field. Figure 1 illustrates how agents are evaluated by their related critics (local and Lead Critics).

Central to HLC is a nested sequential update procedure. For each selected agent, we update its policy sequentially across its related critics, ordered by increasing receptive field, re-sampling the agent’s action after every update so each critic evaluates the latest policy. We also update agents sequentially (random order) while conditioning each update on the most recent policies of the others. This design avoids conflicting gradients that arise when averaging multi-critic signals, preserves stable learning signals, and enables policies to internalize both local and group/global objectives.

To support multi-level guidance, we introduce an actor architecture that fuses multiple processing paths via a mixture-of-experts style module with cross-attention. This yields richer representations aligned with the hierarchy of critics while remaining locally executable at inference.

Empirically, HLC delivers improved performance, high sample efficiency, and robust policies on cooperative, non-communicative, partially observable MARL benchmarks. Across tasks and agent scales, HLC consistently outperforms single-hierarchy baselines—independent (local only) and centralized (global only)—and scales robustly with increasing numbers of agents and task difficulty. Video and code are provided in the supplementary material.

We summarize our contributions as follows:
\begin{itemize}
    \item  We propose the Hierarchical Lead Critic (HLC) framework, which integrates multi-level perspectives (local, group-level, centralized) for cooperative MARL, bridging the gap between single-perspective optimization approaches.
    \item A nested sequential updating scheme that enables stable multi-critic interactions, avoiding gradient conflicts while preserving locality during training.
    \item An new actor model architecture utilizing a mixture-of-experts style module with cross-attention, designed for efficient multi-level coordination.
    \item Introduction of two novel drone benchmarks—Escort and Surveillance—that test agent cooperation, scalability, and performance under partial observability.

\end{itemize}

\section{Related Work}
State-of-the-art MARL methods \cite{MAT, HASAC, SABLE} commonly rely on a single centralized critic to guide the actors during training, while independent versions of single-agent RL methods use local critics and treat other agents as part of the environment. Even though methods that learn centralized and decentralized critics simultaneously exist, they often collapse them into a scalar learning signal as discussed in section~\ref{Related_Work:MultipleCritics}.
Crucially, algorithms that optimize for multiple critic signals directly have not been explored. In the following section we give an overview of MARL algorithms and introduce methods that learn from multiple critics as these areas are the closest related to HLC conceptually.

\subsection{MARL} \label{Related_Work:MARL}
Centralized critics in MARL algorithms like MADDPG~\citep{MADDPG}, COMA~\citep{COMA}, MAAC~\citep{MAAC}, MAT~\citep{MAT}, HASAC~\citep{HASAC}, SABLE~\citep{SABLE} are used to mitigate the non-stationarity introduced by agents learning concurrently.
MADDPG, as extension of DDPG to multiple agents under the CTDE training paradigm demonstrated the performance improvement of applying centralized critics, but fails to address the multi-agent credit assignment problem. It uses a centralized critic for each agent separately and conditions them on non-local information, such as observations and actions from other agents or global state information.
COMA also employs a centralized critic, but tackles the problem of multi-agent credit assignment by subtracting a counterfactual baseline to marginalize out an agent's action.
MAAC successfully applied the attention mechanism to the concept of centralized critics, and MAT introduces a Transformer based on-policy encoder-decoder architecture with a sequential update scheme based on the Multi-Agent Advantage Decomposition theorem proposed in~\citep{MA-AdvantageDecompositionOriginal, MA-AdvantageDecompositionTRPO}. SABLE combines MAT with retention to improve performance. \\
HASAC, as the current SOTA algorithm in multi-agent continuous control, is based on the MEHAML framework, an extension of HAML~\citep{HAML} to maximum entropy Reinforcement Learning (RL). Building on the Multi-Agent Advantage Decomposition theorem, it guarantees theoretical monotonic improvement and derives strong algorithms like HAPPO and HATD3. In HASAC, agents learn a stochastic policy and are updated sequentially in random order, with the centralized critic conditioned on the most recent policy from each agent and a global state. After each policy update, a new action is sampled to guide the update of the next agent.
HLC leverages the state-of-the-art idea of sequential updating between agents and the centralized critic learning style as proposed in HASAC, which is discussed in section~\ref{Section:HLC}.

\subsection{Learning from Multiple Critics}\label{Related_Work:MultipleCritics}
The use of multiple critics with a single actor has mostly been explored in the fields of Multi-Task Reinforcement Learning (MTRL)~\citep{MultiCriticAL, SLIM} and MARL~\citep{AMC-TD3, DMCN, MARL-TeamR-Decomp, MAGAC}.
These approaches typically combine multiple Q-values or gradients using a weighted average to update the actor. While this method offers good empirical performance with low implementation complexity, it requires tuning the weights, which introduces additional hyperparameters that often need extensive domain expertise. Furthermore, gradients from multiple critics can interfere destructively as they are calculated simultaneously, which can destabilize the training process. To address conflicting gradients, solutions have been proposed with techniques like CoNAL~\citep{CoNAL} that mitigate conflicting gradients during the architecture learning process, or PCGrad~\citep{PCGrad} that performs gradient surgery. In MARL, using multiple centralized critics has been explored by MA-POCA~\citep{MA-POCA}, where a task is split into two new simpler tasks. Each task has a centralized critic that coordinates a subgroup of agents, without overlap between the groups, with each subgroup receiving task-specific observations and rewards. MA-POCA demonstrated that dividing knowledge among multiple critics can improve learning performance. In contrast, HLC learns from multiple critics per actor, which allows optimization from multiple hierarchies. With the HLC sequential updating scheme, after each update a new action is inferenced and evaluated by the next critic. This mirrors the basic idea in actor-critic that critic targets should reflect the most recent policy to maintain stable gradients. Thus, HLC avoids the issues of conflicting gradients and the limitations of weighted averaging.

\section{Preliminaries}
We consider the cooperative setting of Partially Observable Markov Games (POMG)~\citep{MarkovGames} as generalization of Decentralized Partially Observable Markov Decision Processes (Dec-POMDP) with individual reward functions. A POMG is described by the tuple (\(N, S, \{O^i\}, A, \{R^i\}, P, \gamma\)), where \(N = \{1,...n\}\) denotes the set of \(n\) agents. \(S\) refers to the finite set of states and \(\{A^i\}\) denotes the action space of agent \(i \in N\), resulting in the joint action space \(A\). Agents only observe their local observation \(o^i_t \in O^i\) instead of the full state, where \(O^i\) is the observation space of agent \(i\).  Given a state \(s_t \in S\) and joint action \(a_t \in A\), the environment transitions to a new state \(s_{t+1} \in S\) by the transition probability function \(P(s_{t+1} \mid s_t,a_t): S \times A \times S \rightarrow [0,1]\) and returns reward signals \(r_t^i \in R^i: S \times A \times S \rightarrow \mathbb{R}\). \(\gamma \in [0,1)\) is the discount factor.\\
HLC is trained under the CTDE paradigm, allowing access to non-local information during training, whereas at execution time the policies operate using only local observations.

\section{Hierarchical Lead Critic} \label{Section:HLC}
In this section, we introduce the Hierarchical Lead Critic - a novel sequential updating scheme and architecture for learning from multiple hierarchy levels with actor-critic algorithms.
\subsection{Lead Critics}
We define a Lead Critic as one that evaluates a group of agents. In contrast, a centralized critic always evaluates all agents and a local critic only evaluates its agent. The centralized and local critic are therefore extreme cases of the Lead Critic concept. Multiple Lead Critics can be organized in parallel or hierarchically to group agents based on principles like locality or functionality. Each Lead Critic incorporates a non-local reward, which is either a distinct group reward or a combination of local rewards, to encourage coordination towards a shared goal. We propose implementing the Lead Critic as a Transformer-Encoder based architecture, optionally incorporating a feature extractor for observation-action pairs. This design allows handling heterogeneous agents and environments without a global state; instead, we rely on observation-action pairs from agents. A visualization is provided in \autoref{Appendix:TransformerEncoder}.
\subsection{Sequential Updates}
\begin{figure}[ht]
  \vskip 0.2in
  \begin{center}
    \centerline{\includegraphics[width=\columnwidth]{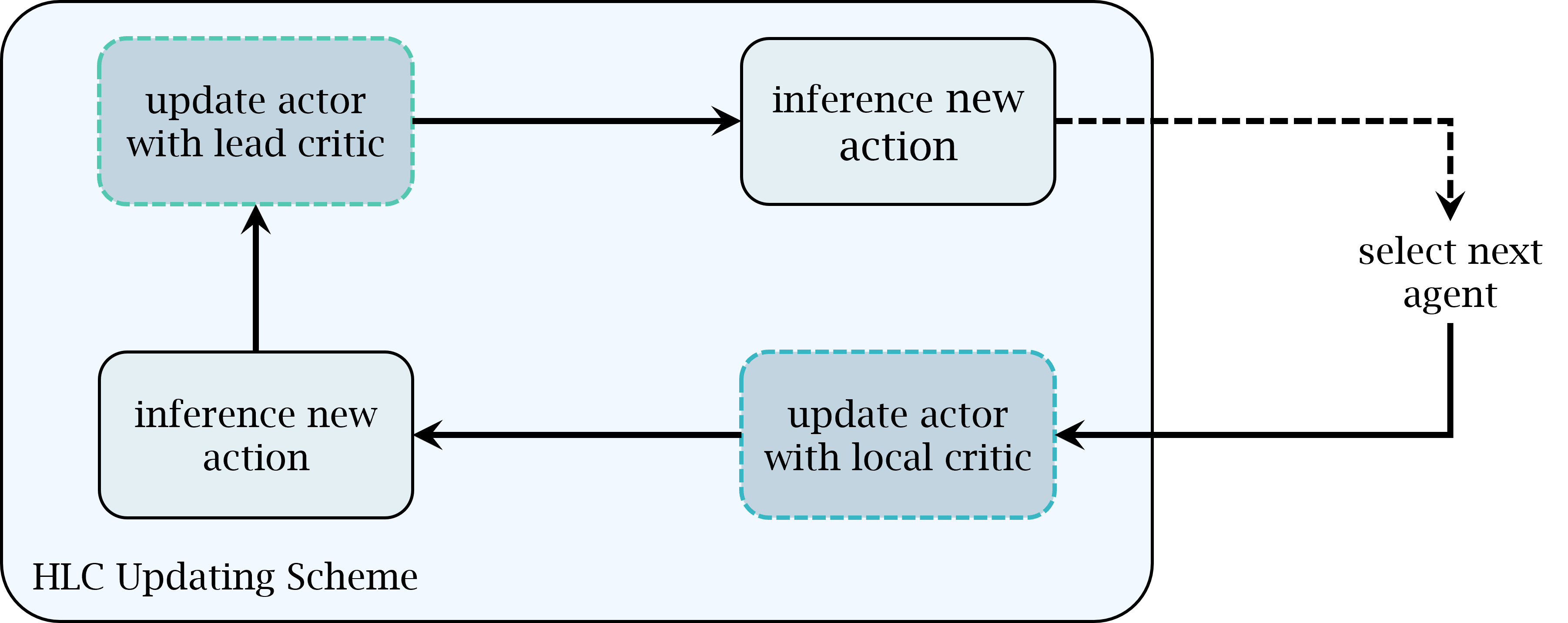}}
    \caption{
      HLC sequential updating scheme for a selected agent with a local critic and a Lead Critic.
    }
    \label{fig:HLC-Sequential}
  \end{center}
\end{figure}

With HLC, we introduce a novel sequential updating scheme between multiple critics when updating a single actor, as shown in Figure~\ref{fig:HLC-Sequential}. The actor is updated by multiple critics sequentially, ordered by the size of their receptive fields, which refers to the number of agents they evaluate. After each policy update, we inference a new action, ensuring that each critic always evaluates the latest policy. This avoids the problem of conflicting gradients, underlined by a separate discussion comparing our approach with a single averaged update found in \autoref{Appendix:SequentialUpdates}. After an agent has been updated by all relevant critics, the next agent is sequentially updated and as the last update of an agent always focuses on the most cooperative behavior (critic with largest receptive field), the next agent’s update can rely on the most cooperative version of other agents. This second sequential updating scheme, between agents, follows~\cite{HASAC}. The HLC update procedure, consisting of two sequential updating schemes, allows agents to act according to multiple hierarchies, while promoting cooperation. Note that~\cite{HASAC} can be recovered from HLC by restricting the setup to a single centralized critic.

\subsection{HLC Actor}
\begin{figure}[ht]
  \vskip 0.2in
  \begin{center}
    \centerline{\includegraphics[width=\columnwidth]{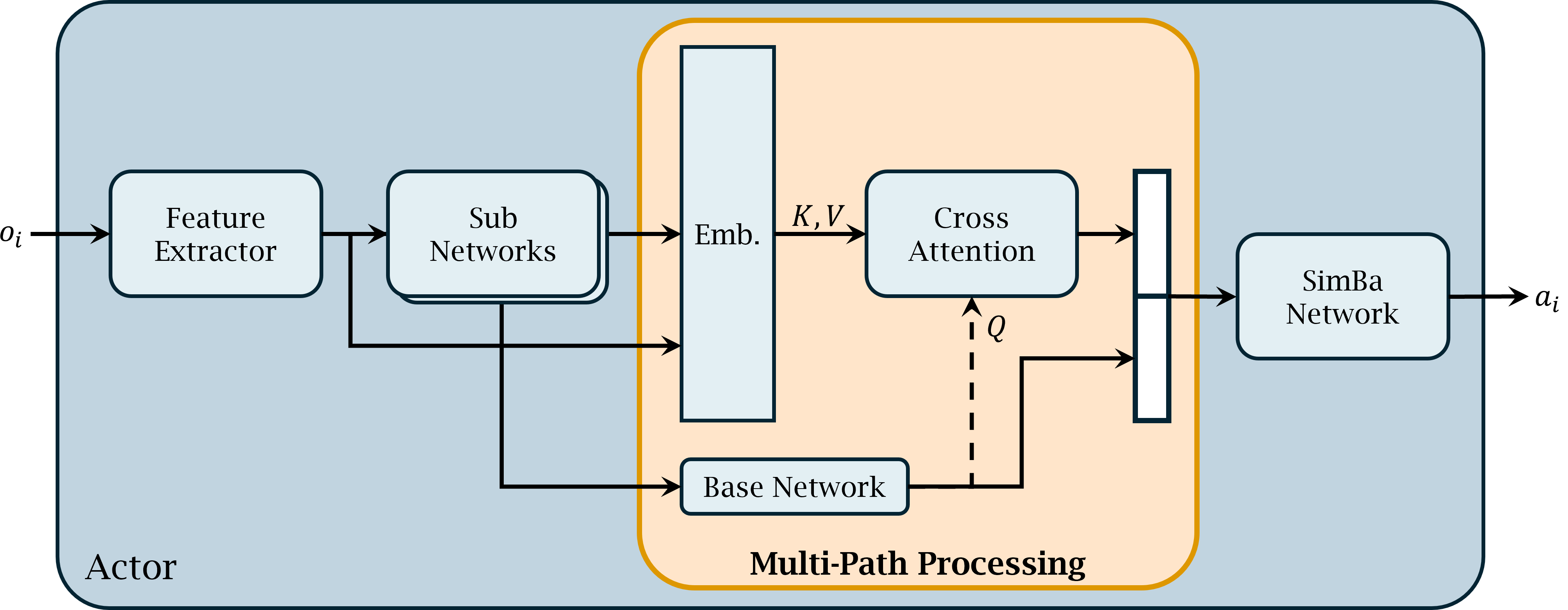}}
    \caption{
      HLC actor. The observation is processed through multiple paths that are combined through cross-attention and concatenation.
    }
    \label{fig:HLC-Actor}
  \end{center}
\end{figure}

We also propose a novel actor architecture, tailored for multi hierarchy optimization, where we combine multiple processing paths through the actor with parallel mixture-of-experts style subnetworks and cross-attention. The architecture is shown in Figure~\ref{fig:HLC-Actor}. After a common feature extractor, multiple distinct subnetworks compute different representations of the input. The subnetwork structure in this mixture-of-experts style module is inspired structurally by mixture-of-encoders from MTRL ~\citep{CARE, AMESAC,MOORE}. Note, that it differs by being constrained to local observations of the respective agent and the application in the context of MARL, centralized critics and our sequential update scheme.\\
The subnetwork outputs are embedded and processed with cross attention. The embedding consists of a shared dense layer, mapping to the same latent space, sinusoidal positional encoding and layer norm. We further process the cross-attention output with an inverted bottleneck MLP with ReLU activations, where the hidden dimension is enlarged by a factor of 4. In parallel to this attention path, the base network path provides general processing of the observations and generates the query for the cross-attention. The outputs from the base network path and cross-attention path are concatenated and processed by a SimBa Net~\cite{SimBa} for action prediction.\\
During backpropagation the base network can be effectively updated through the residual SimBa connection. As we detach the query, there is no gradient from the attention path to the base network, decoupling the base networks learning process. This specific setup allows the HLC actor to act in early training mainly based on the base network and SimBa residual connection, while the cross-attention path and SimBa Net learn to provide more nuanced information for complex emergent cooperative behavior towards later training stages.

\subsection{HLC with SAC}
We explore SAC~\cite{SAC, SAC-Applications} as the underlying actor-critic algorithm for HLC. With a Lead Critic, we learn group Q-values \(Q_{LC{\psi_{i \in \{1, 2\}}}}\) that are conditioned on observation-action pairs \(o^{1:m}, a^{1:m}\), for a group of \(m\) agents. Details on learning the group Q-values are given in \autoref{Appendix:HLC-SAC}. For calculating the actor loss \(J_{\pi^k}\) of agent \(k\) from a Lead Critic, we use its entropy coefficient \(\alpha^k\) and use the average log probability of agents associated with the current Lead Critic for calculating the entropy:

\begin{equation}\label{eq:LC_actor_loss}
\begin{aligned}
J_{\pi^k}(\theta^k) = \mathbb{E}_{\substack{o_{t}^{1:m} \sim D \\ \text{ }a_{{t}}^{1:m} \sim \pi_{\theta^{1:m}}^{1:m}}} 
\left[
H^k_{LC}
-\min_{j \in \{1, 2\}}  Q_{LC_{\psi_j}}(o_{t}^{1:m}, a_{t}^{1:m})
\right] \\
\text{with } H^k_{LC} = \alpha^k\cdot\left[\frac{1}{m}\sum_{l=1}^m{\log \pi^l_{\theta^l}(a_{{t}}^{l} | o_{{t}}^{l})}\right]
\end{aligned}
\end{equation}

Note that all actions originate from the current policies, and gradients are only calculated for agent \(k\). We note that the sequential update scheme for updating agents related to a certain Lead Critic follows~\citep{HASAC} by conditioning each actor's loss calculation on the most recent policies. \\
In Algorithm \ref{alg:HLCFullPseudocode}, we show a training step for HLC and denote statements that apply SAC calculations as (SAC).

\begin{algorithm}[thp]
\caption{Hierarchical Lead Critic - Training step}
\label{alg:HLCFullPseudocode}
\begin{algorithmic}
    % \ttfamily
    \STATE\textbf{Input:} Gradient steps G, number of agents n, agent update interval: F, policies \(\pi_{\theta^{1:n}}^{1:n}\), local critics \(Q^{1:n}_{\phi_{i \in \{1,2\}}^{1:n}}\) with target networks \(Q^{1:n}_{\bar\phi_{i \in \{1,2\}}^{1:n}}\), lead critics \(Q^{1:m}_{LC{\psi_{i \in \{1, 2\}}}}\) with target networks \(Q^{1:m}_{LC{\bar\psi_{i \in \{1, 2\}}}}\), replay buffer D;
    \FORALL{gradient step \(g\) = 1,..., \(G\)}
        \STATE Sample minibatch \(B \sim D\);
        \STATE Select actions \(a_{{t}}^{1:n} \sim \pi^{1:n}_{\theta^{1:n}}(\cdot | o_{{t}}^{1:n})\)
        \STATE Update lead critics (\autoref{eq:LC_Critic_Update});
        \IF{$g\% F\neq 0$}
            \STATE \textbf{continue}
        \ENDIF
        \STATE
        \STATE Draw random permutation \(P\) of agents \(i_{1:n}\);
        \FORALL{\(k \in P\)}
            \STATE Update local critic \(Q^k\) with \(a_{{t}}^{k}\) (SAC);
            \STATE Update policy \(\pi^k_{\theta^k}\) with local critic \(Q^k\) (SAC);
            \STATE Get lead critics \(LC_k\) associated with agent k;
            \FORALL{\(Q^k_{LC} \in LC_k\)}
                \STATE Select action \(a_{{t}}^{k} \sim \pi^k_{\theta^k}(\cdot | o_{{t}}^{k})\)
                \STATE Update policy \(\pi^k_{\theta^k}\) with \(Q^k_{LC}\) (\autoref{eq:LC_actor_loss});
            \ENDFOR
            \STATE Select action \(a_{{t}}^{k} \sim \pi^k_{\theta^k}(\cdot | o_{{t}}^{k})\)
            \STATE Update entropy of agent k (SAC);
        \ENDFOR
        \STATE Update target networks with polyak averaging;
    \ENDFOR
\end{algorithmic}
\end{algorithm}

\section{Experiments}\label{sec:result}
\begin{figure}[h]
\begin{centering}
    \subfloat[SimpleSpread]
        {\includegraphics[scale=0.5]{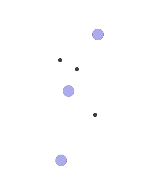}\label{subfloat:SimpleSpread}}
    \subfloat[Escort8]
    {
    \includegraphics[scale=0.25]{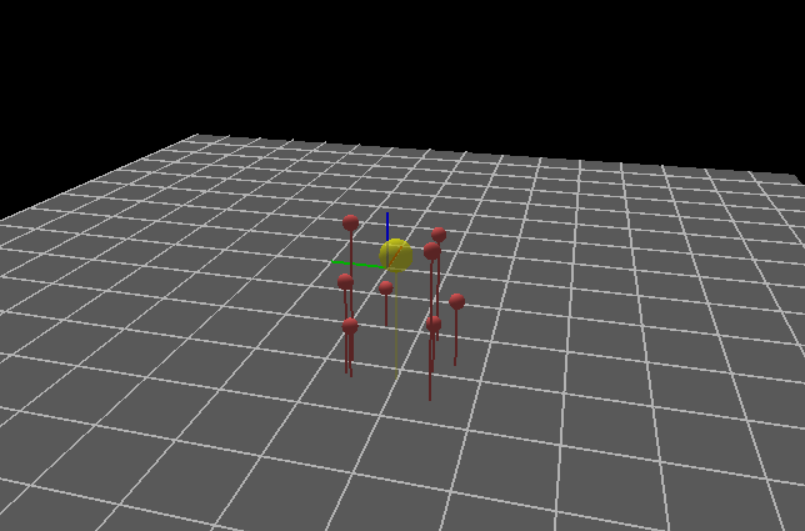}
    \label{subfloat:Escort8}
    }\\
    \subfloat[Surveillance4]
    {
    \includegraphics[scale=0.45]{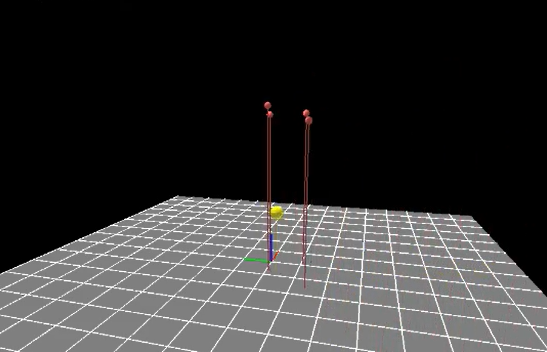}
    \label{subfloat:Surveillance4}
    }
        
    \caption{
    SimpleSpread \protect\subref{subfloat:SimpleSpread}~\cite{MPE_Pettingzoo}, Momaland Escort \protect\subref{subfloat:Escort8}, and Surveillance \protect\subref{subfloat:Surveillance4} environments used in our evaluations. In SimpleSpread, black circles are landmarks and purple particles are agents. In Escort8, agent drones (red) escort a target (yellow), while keeping an implicit formation. In Surveillance4, agents observe a target with fixed relative altitude and distribute themselves evenly around the target.
    }
    \label{fig:EvaluationEnvs}
\end{centering}
\end{figure}

In this section, we evaluate the effectiveness of our HLC approach under the setting of cooperative tasks within CTDE, different types of partial observability, and no communication between the agents.
The environments used in our evaluations are visualized in Figure~\ref{fig:EvaluationEnvs}.
We conduct experiments on an extended version of the MOMAland Escort task~\citep{felten2024momaland} and a novel Surveillance task, described in section~\ref{sec:MomalandTasks}, and on the pettingzoo~\citep{MPE_Pettingzoo} version of the MPE~\citep{MPE} environment SimpleSpread.
% BASELINES - HASAC
We compare HLC with the current SOTA in off-policy MARL - HASAC~\citep{HASAC} and Independent-SAC (ISAC). HASAC and ISAC can be interpreted as extreme variants of the HLC concept, with either a single centralized critic or only local critics. Therefore, comparing with these baselines offers direct insights into the effectiveness of HLC that learns directly from both local and centralized critics.

Training details and hyperparameters can be found in \autoref{Appendix:ExperimentalDetails}.

\subsection{MOMAland Escort \& Surveillance}\label{sec:MomalandTasks}
We use the Escort task from MOMAland~\citep{felten2024momaland} version 0.1.1, where simple drones escort a moving target. We extend this version of the Escort task to a fully functional benchmark environment, by introducing proper seeding and randomization, a new scaled reward function and more complex target movement. The code, containing all changes, is available in the supplementary material.\\
The observation of each agent is generally constructed by relative positions of the target and other agents. The order is fixed and all agents are observed. Additionally, agents observe their own height and we stack the current and last observation to allow agents to infer movement directions. As partial observability describes the incomplete observation of a state, we consider two types of partial observability in our experiments: relative positions and limiting observations to neighbors. The former is used in the general case as described and the latter additionally limits the observations of other agents to a subset of closest neighbors (results in section~\ref{sec:AblationStudy} - Escort8 variants). The action space for each agent is a three-dimensional vector with entries \([-1,1]\) and actions are applied with a factor of \(0.2\) directly to the current position of an agent.\\
We build the reward generally from multiple components, each representing a desired behavior. For the Escort task, the reward consists of two components \(r_{distance}\) and \(r_{formation}\) that benefit agents closely following the moving target, while keeping a stable formation. We denote the distance between two entities a and b as \(d(a,b)\), the current agent as agent \(i\) and the average position of all agents as center of mass \(COM\).
The reward component \(r_{distance}\) incentivizes the agents to move closer to the target and keep the targets' position close to the center of mass of all agents.
% Single agent reward component DISTANCE
\begin{equation}\label{eq:Escort_VectorReward}
\begin{aligned}
r^i_{distance} = - \frac{1}{2}\left(d(i,target) + d(COM, target)\right)
\end{aligned}
\end{equation}
The component \(r_{formation}\) embodies the cooperative formation objective by rewarding agents for staying at a distance of 1 towards the three closest surrounding agents. This formulation allows scalability to larger agent quantities, as in swarm like behavior.

% Single agent reward component FORMATION
\begin{equation}\label{eq:Escort_VectorReward}
\begin{aligned}
r^i_{formation} = - \frac{1}{2}\left(|mean\_d-1| + |max\_d-1|\right) \\
mean\_d = \frac{1}{3}\sum_{j=1}^{n_{closest3}}{d(i,j}) \\
max\_d = max_{j \in n_{closest3}} d(i,j) \\
\end{aligned}
\end{equation}

The components are normalized by the map size and clipped between \([-1, 0]\), while the average of these components is used to create a scalar reward \(r_{scalar}\) for each agent, respectively. A team reward is then calculated as the sum of all scalar rewards:

\begin{equation}\label{eq:Escort_ScalarReward}
\begin{aligned}
r^i_{scalar} =  0.5 \cdot r^i_{distance} + 0.5 \cdot r^i_{formation}
\end{aligned}
\end{equation}

\begin{equation}\label{eq:Escort_TeamReward}
\begin{aligned}
r_{team} = \sum_{i=0}^n{r^i_{scalar}}
\end{aligned}
\end{equation}

\begin{table}[t]
  \caption{Termination rewards in Escort \& Surveillance tasks}
  \label{sample-table}
  \begin{center}
    \begin{small}
      \begin{sc}
        \begin{tabular}{lc}
          \toprule
          Task  & Termination reward  \\
          \midrule
          Escort3 & -200  \\
          Escort &  -200 \\
          Escort8 & -500 \\
          Surveillance4 & -200  \\
          \bottomrule
        \end{tabular}
      \end{sc}
    \end{small}
  \end{center}
  \vskip -0.1in
  \label{tab:MomalandTerminationReward}
\end{table}

The episodes terminates early if agent collides with the ground or other agents. The termination rewards in Table~\ref{tab:MomalandTerminationReward} then override the scalar reward signal of the colliding agent. Note that with maximum 200 steps per episode (without early termination), an agent receives at worst an episodic return of \(-200\) when finishing the episode without early termination. \\
In addition, we also introduce a novel Surveillance task, where an agent \(i\) gets the rewards \(r^i_{height}\) for keeping a fixed relative altitude to the target, \(r^i_{torus}\) to distribute agents evenly around the target within a specified distance interval, and \(r^i_{relH}\) to maintain the same altitude as other agents. We denote the agent altitude as \(z^i\), the target altitude as \(z^t\), the average altitude of all agents as \(z_{avg}\) and the xy-distance between to entities a and b as \(d_{xy}(a,b)\). Agents also receive the reward \(r^i_{formation}\) as introduced in the Escort task in \autoref{eq:Escort_VectorReward}.

\begin{equation}\label{eq:Surv_HeightReward}
\begin{aligned}
r^i_{height} = -|(z^i - z^t) - 4| \\
d = d_{xy}(i,t) \\
r^i_{torus} = -
\begin{cases}
1.0 - d & \text{if } d < 1.0 \\
d - 2 & \text{if } d > 2.0 \\
0 & \text{else}
\end{cases}
\\
z_d = |z^i - z_{avg}|\\
r^i_{relH} = 
\begin{cases}
-z_d   & \text{if } z_d > 0.1 \\
0
\end{cases}
\end{aligned}
\end{equation}
Following the Escort task, we normalize each component by the map size and clip them between -1 and 0. Also on collision, agents only receive the termination reward. From these reward components we calculate the two objectives:
\begin{equation}\label{eq:Surv_HeightReward}
\begin{aligned}
r^i_{local} = 0.5 \cdot r^i_{height} + 0.5 \cdot r^i_{torus} \\
r^i_{cooperation} = 0.5 \cdot r^i_{relH} + 0.5 \cdot r^i_{formation}
\end{aligned}
\end{equation}

As in the Escort task, compare~\autoref{eq:Escort_ScalarReward}, a scalar reward is calculated as average of \(r^i_{local}\) and \(r^i_{cooperation}\) and the team reward is calculated with \autoref{eq:Escort_TeamReward}.\\
A detailed description of general changes to MOMAland is given in \autoref{Appendix:ExperimentalDetails}.

\subsection{Experimental Results}
\begin{figure*}[t]
\begin{centering}
    \subfloat[Escort3]
        {\includegraphics[scale=0.22]{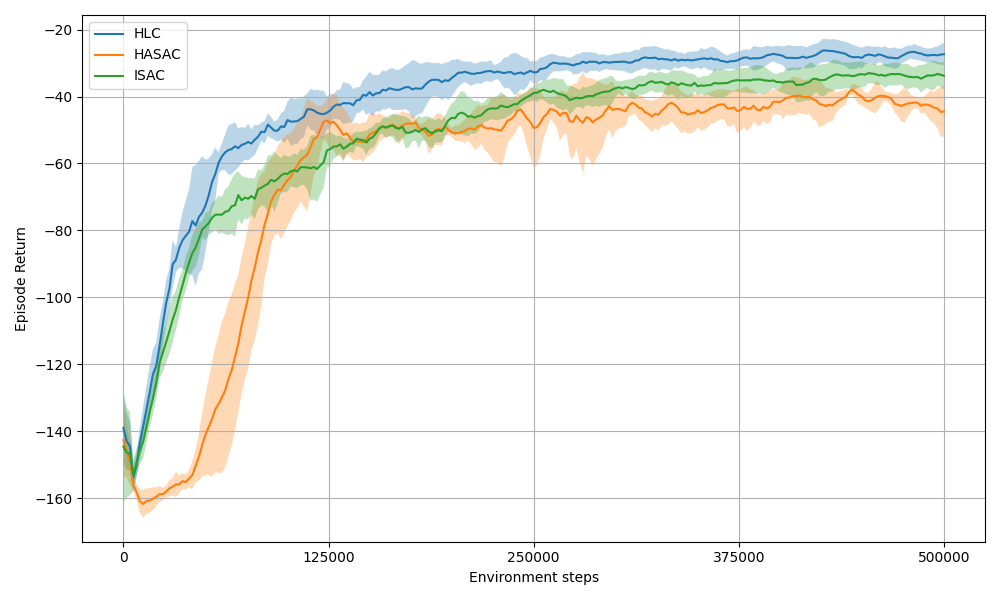}\label{subfloat:Escort3_MainResult}}
    \subfloat[Escort8]
        {\includegraphics[scale=0.22]{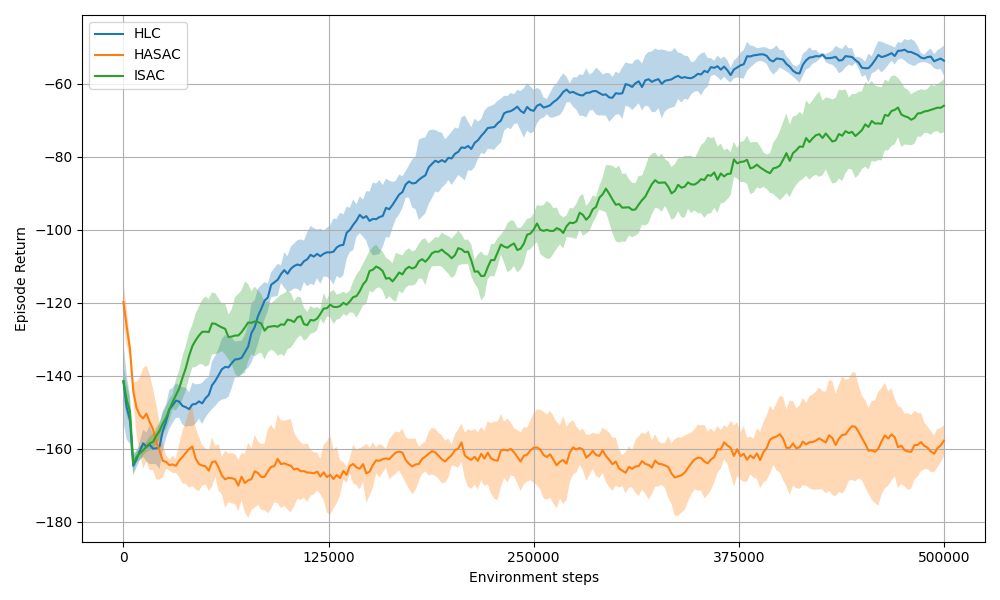}\label{subfloat:Escort8_MainResult}}
    \subfloat[Surveillance4]
        {\includegraphics[scale=0.22]{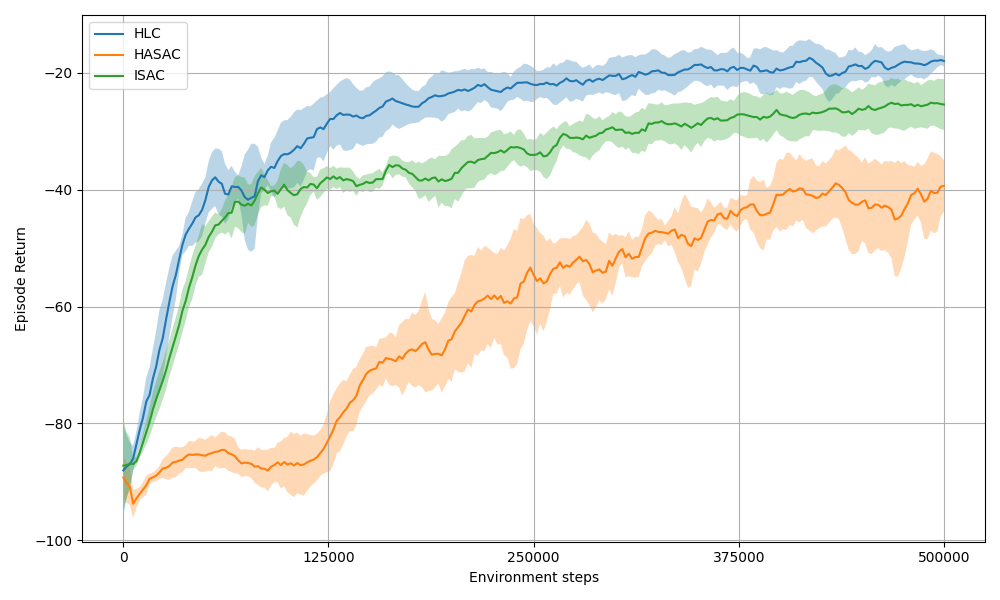}\label{subfloat:Surveillance4_MainResult}}\\
    \subfloat[Escort3 - Episode length]
        {\includegraphics[scale=0.22]{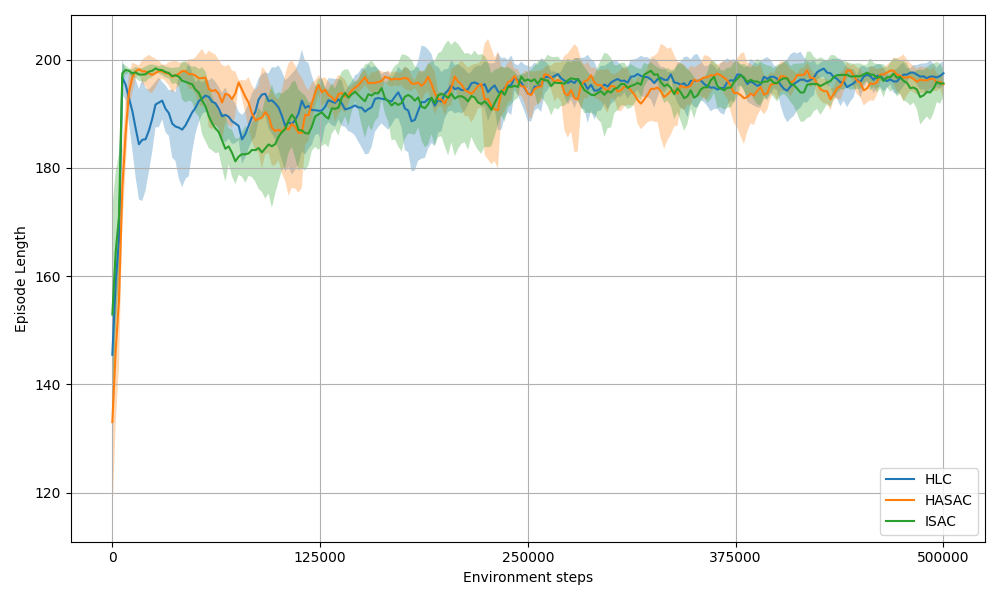}\label{subfloat:Escort3_200_length_MainResult}}
    \subfloat[Escort8 - Episode length]
        {\includegraphics[scale=0.22]{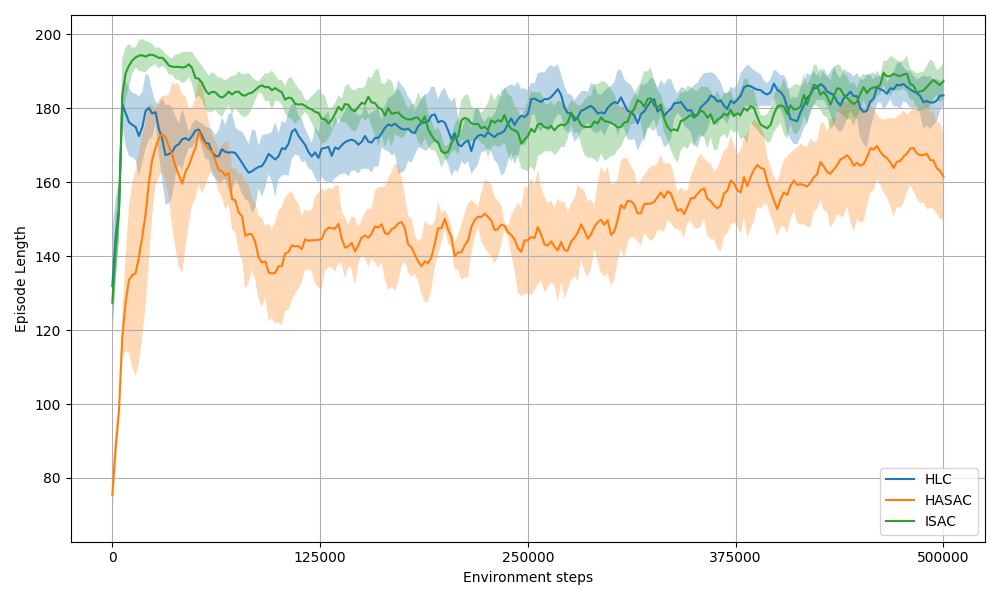}\label{subfloat:Escort8_200_length_MainResult}}
    \subfloat[SimpleSpread]
        {\includegraphics[scale=0.22]{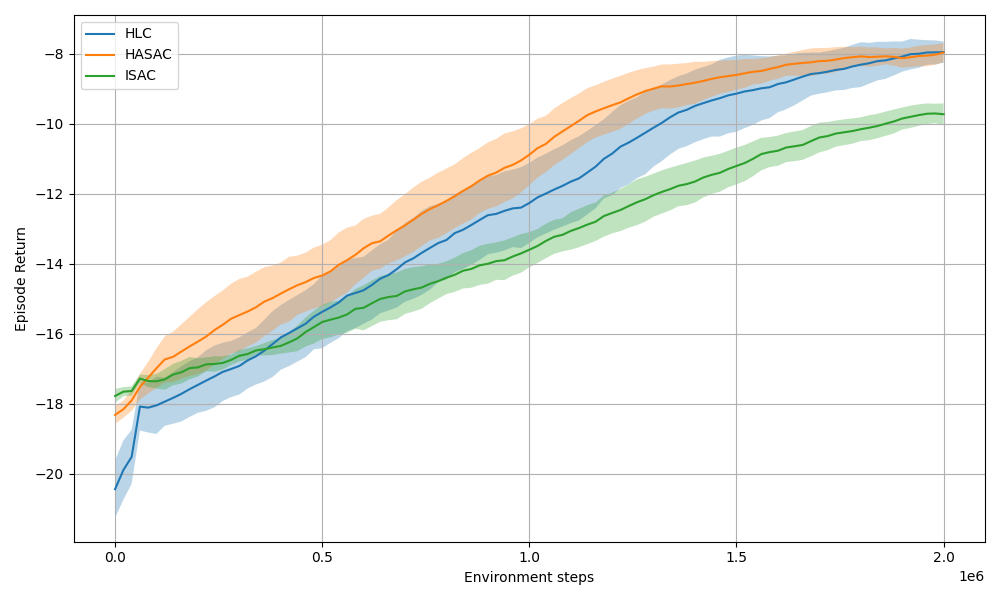}\label{fig:SimpleSpread_MainResult}}
    \caption{
    Performance of HLC compared to HASAC and ISAC on the Escort tasks with episode lengths, Surveillance task and SimpleSpread.
    }
    \label{fig:Main_Results}
\end{centering}
\end{figure*}
\textbf{Escort3 and Escort8}\\
On the Escort tasks, HLC generally outperforms both HASAC and ISAC in final performance and sample efficiency (see Figures~\ref{subfloat:Escort3_MainResult} and~\ref{subfloat:Escort8_MainResult}). The strong performance of ISAC demonstrates that local rewards are important to progress in these tasks. On the other hand, HASAC struggles on Escort3 and degrades to termination avoidance on Escort8, showcasing that solely relying on a global reward can be troublesome (episode lengths are provided in Figure~\ref{subfloat:Escort3_200_length_MainResult} and~\ref{subfloat:Escort8_200_length_MainResult}). HLC manages to combine the local reward benefits with improved cooperative behavior to achieve unmatched performance. It effectively integrates the global reward without degrading the performance as seen with HASAC.\\
We provide ablations on the death penalty and partial observability settings of the Escort8 task in section \ref{sec:AblationStudy}. 

\textbf{Surveillance4}\\
On the Surveillance task, a similar pattern to the Escort tasks can be observed. A good individual performance of an agent is vital to an effective performance of the group. HLC manages again to leverage the individual and global components to achieve unmatched performance and sample efficiency, while ISAC offers limited performance and HASAC struggles to solve the task.

\textbf{Simple Spread}\\
On SimpleSpread agents are mainly rewarded based on the global objective, while local penalties are introduced for collisions. Despite this setup, HLC exhibits competitive performance compared to HASAC as seen in Figure~\ref{fig:SimpleSpread_MainResult}. This shows that HLC can also handle sparse local reward setups and not degrade to a worse performance as seen with ISAC.

\subsection{Ablation Study}\label{sec:AblationStudy}
\begin{figure*}[tb]
\begin{centering}
    \subfloat[Escort8 - Death 320]
        {\includegraphics[scale=0.22]{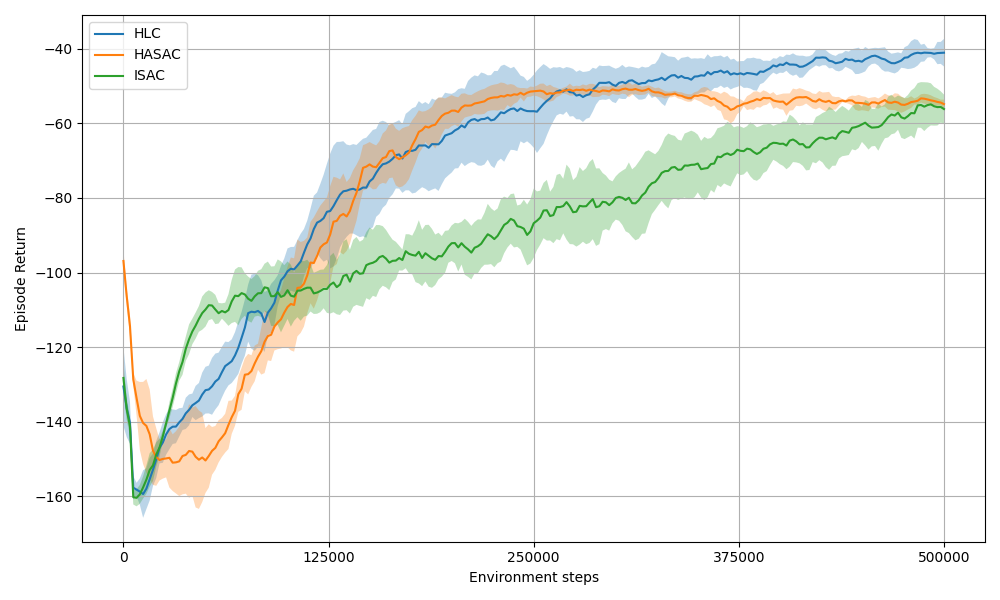}\label{subfloat:Escort8_320_Main_Result}}
    \subfloat[Escort8 - HLC-Simple]
        {\includegraphics[scale=0.22]{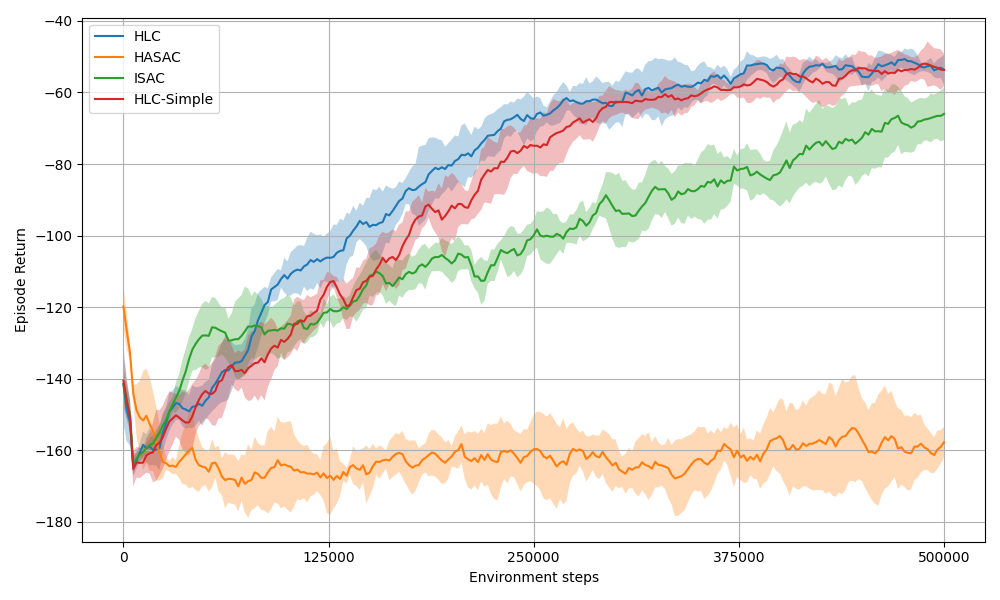}\label{subfloat:AblationStudy_Escort8_500_HLC-Simple}}
    \subfloat[Escort8 - Partial 2]
        {\includegraphics[scale=0.22]{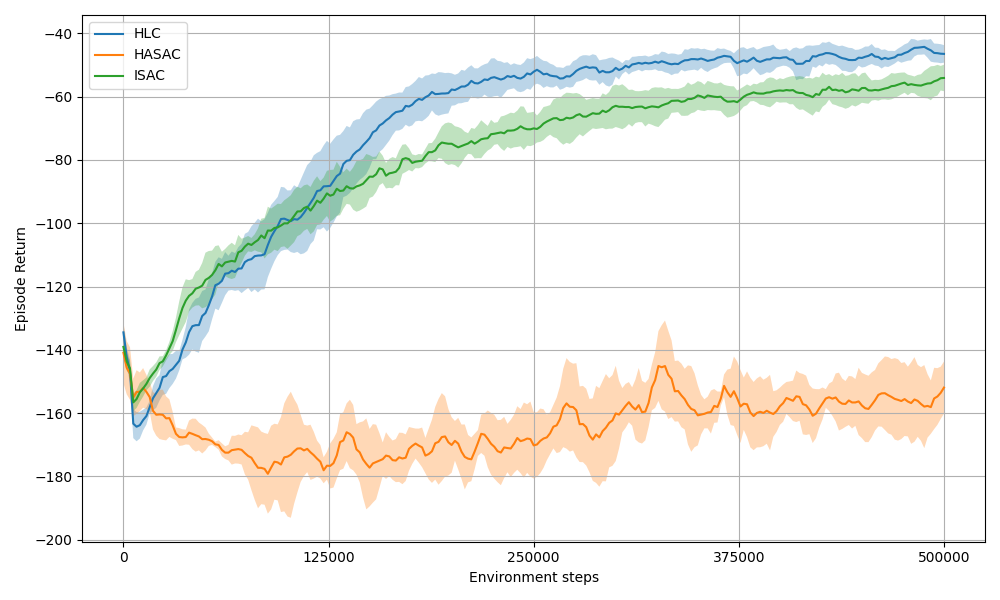}\label{subfloat:Escort8_Partial2}}\\
    \subfloat[Escort8 - Death 320 - Episode length]
        {\includegraphics[scale=0.22]{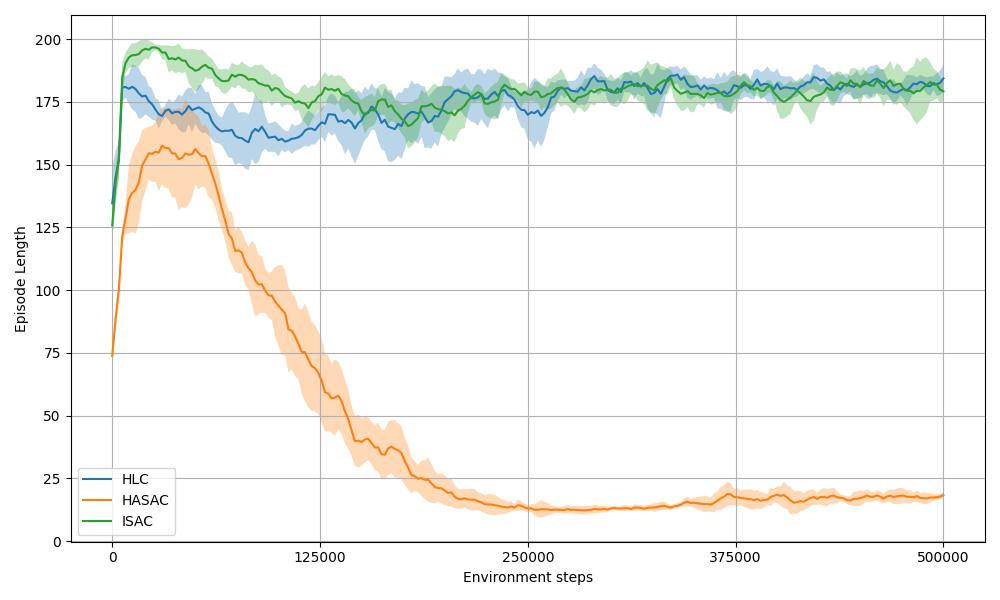}\label{subfloat:Escort8_320_length_Main_Result}}
    \subfloat[Escort8 - SingleLogp]
        {\includegraphics[scale=0.22]{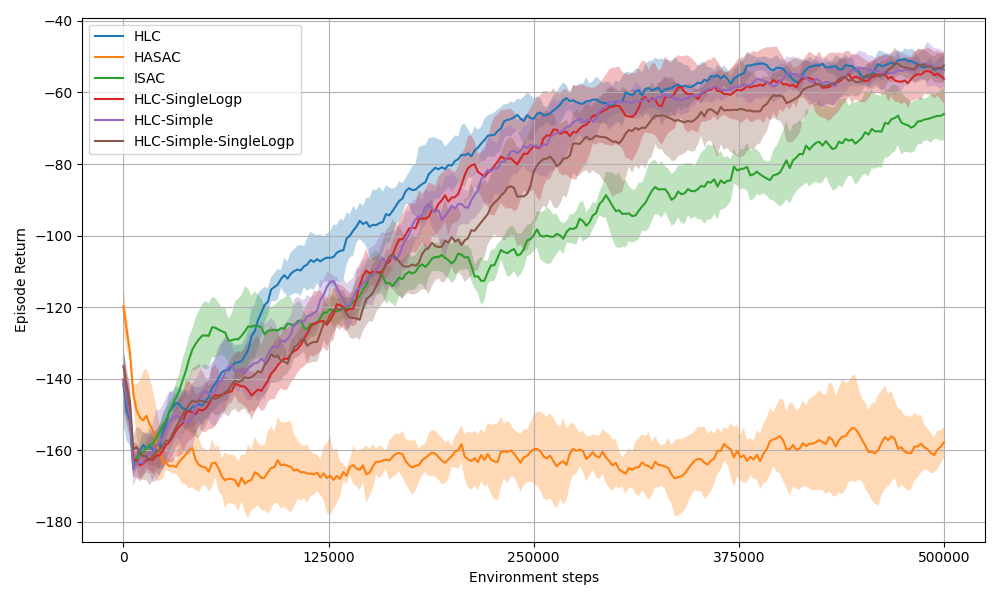}\label{subfloat:AblationStudy_HLC-Simple-SingleLogp}}
    \subfloat[Escort8 - Partial 4]
        {\includegraphics[scale=0.22]{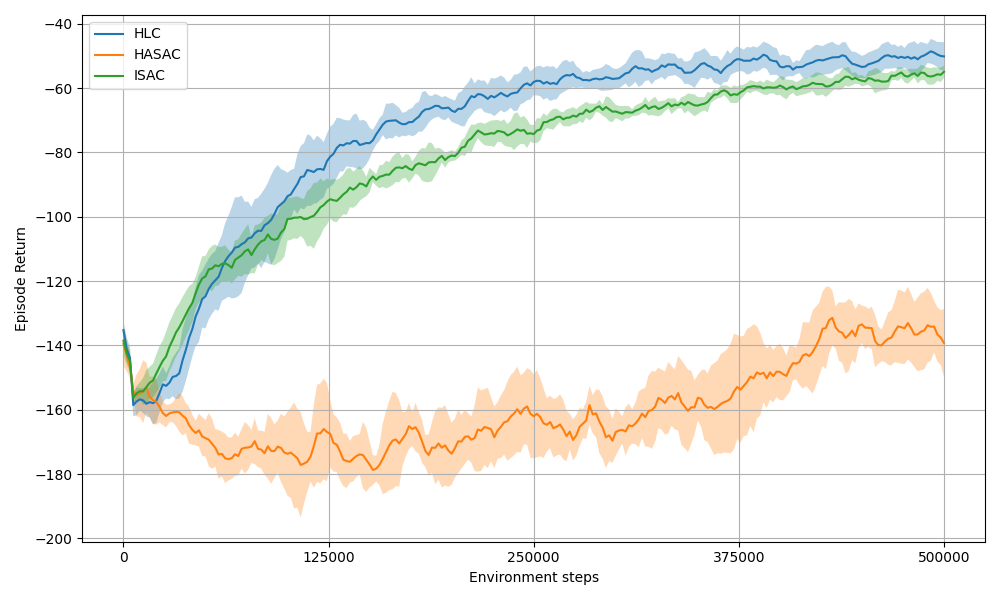}\label{fig:Escort8_Partial4}}
    \caption{
    Ablation Study on HLC components and Escort 8 variants.
    }
    \label{fig:Main_Results}
\end{centering}
\end{figure*}
\textbf{HLC components}\\
We investigate different components of HLC, consisting of the actor architecture and the averaged log probability (see \autoref{eq:LC_actor_loss}).\\
For the architecture, we evaluate HLC-Simple. A variant, where we swap the HLC actor with a simple 2-layer MLP with ReLU activations, mirroring the HASAC and ISAC actor setup. We report the results in Figure~\ref{subfloat:AblationStudy_Escort8_500_HLC-Simple}. HLC-Simple outperforms HASAC and ISAC, showcasing that the proposed sequential updating scheme and hierarchical setup of HLC increase the performance already by themselves and verifies the effectiveness of leveraging both local and centralized rewards during training. Only the full HLC setup, with advanced actor architecture, achieves better performance.\\
For the averaged log probability, we introduce the Single Logp variants, that employ the HASAC setup of only using each agents' log probability in the actor loss. We report the results for HLC-SingleLogp and HLC-Simple-SingleLogp in Figure~\ref{subfloat:AblationStudy_HLC-Simple-SingleLogp}. A consistent trend emerges, HLC-Simple outperforms HLC-Simple-SingleLogp and HLC outperforms HLC-SingleLogp. The averaged log probability setup produces better performance in both variants consistently.\\
The best performance is still achieved by the full HLC setup, combining the proposed sequential updating scheme, the HLC-actor and the averaged log probabilities.

\textbf{Escort8 variants}\\
We provide additional results on two variants of the Escort8 task to evaluate the robustness of HLC and the baselines to partial observability and a difficult early termination optima.\\
For the partial observability setup, we reduce the observations of each agent to only the relative positions of the closest agents and the relative target position. We report results for agents only observing the two and four closest agents respectively. Also under partial observability HLC outperforms the baselines.\\
To increase the difficulty in respect to early termination optima we purposely reduce the death penalty on termination from \(500\) to \(320\) in the Escort8 task. This results in a difficult local optimum, where agents can try to collide as soon as possible to force early termination. Imagine a single agent colliding with the ground in the first step of an episode. This directly terminates the episode and results in a global reward of roughly \(-320\), including other agents contributions, which are close to zero. Such a global reward can be falsely attributed to all agents evenly achieving a comparably good episodic return of \(-40\) each. Especially in early training a strategy of converging to the local optimum of instant termination can be favorable for the agents.\\
We report the performance in Figure~\ref{subfloat:Escort8_320_Main_Result} and episode length in Figure~\ref{subfloat:Escort8_320_length_Main_Result}. Notably, HASAC quickly falls into the local optimum of early termination, while HLC retains its strong performance and is not negatively impacted due to its reliance on local and global rewards.

\textbf{Additional results}\\
We provide additional results and ablations in \autoref{Appendix:AdditionalResults}, consisting of different HASAC configurations on Escort3, results on Escort5 and results for increasing gradient steps on HASAC and ISAC in Escort8 to investigate if the sample efficiency of the baseline models can easily be increased. To summarize, increasing the gradient steps on HASAC and ISAC does not translate to an improvement in performance, which shows that HLC has the unique advantage under the investigated algorithms of translating additional gradient updates into an increased sample efficiency. On the Escort5 task HLC again outperforms the baselines showcasing its capability of handling different group sizes effectively.\\
We provide a video in the supplementary material, showcasing the performance of HLC on the Escort3, Escort8 and Surveillance4 tasks as detailed in~\autoref{Appendix:VideoResults}.

\section{Conclusion} \label{sec:conclusion}
In this work, we propose HLC as novel actor-critic MARL algorithm that demonstrates the benefits of explicitly learning from multiple perspectives on different hierarchy levels. We demonstrate superior performance and sample efficiency of HLC on cooperative, non-communicative MARL tasks with partial observability. With HLC, we introduce a novel sequential updating scheme between critics that guide the agent to multi hierarchy optimization. In combination with HLC, we propose a novel actor architecture,  which encodes observations in mixture-of-experts style subnetworks and uses multiple paths for acting effectively.\\
Furthermore, we introduce the Lead Critic as concept of critics that evaluate groups of agents. A Lead Critic evaluating a single agent can be interpreted as a local critic, while a Lead Critic evaluating all agents is a centralized critic. The proposed Lead Critic Transformer-Encoder based architecture focuses on effectively combining local observations and actions from a group of agents to guide the group of agents during learning. With the general design of HLC, it can be applied to heterogeneous agents, cooperative environments with partial observability and does not require access to communication between agents. It enables agents to focus on local optimization, while fostering global (or group-level) cooperation. HLC performs competitive against the current SOTA on a common MARL benchmark and outperforms the SOTA on the novel Escort and Surveillance tasks, while displaying high sample efficiency and robustness to partial observability without communication. It shows scalability to larger number of agents, while keeping stability and performance.\\
While HLC outperforms the baselines, the sequential updating scheme increases the training time as the inference time remains comparable. Future works include ideas of parameter sharing between agents and partial parameter sharing between subnetworks to reduce the computational resources. Additionally, exploring the applicability of HLC in Multi-Objective MARL might have relevant potential.

% % Acknowledgements should only appear in the accepted version.
\section*{Acknowledgements}
This research was partially funded by the Bavarian State Ministry for Economic Affairs, Regional Development and Energy under Grant DIK0281.

\section*{Impact Statement}

% Authors are \textbf{required} to include a statement of the potential broader
% impact of their work, including its ethical aspects and future societal
% consequences. This statement should be in an unnumbered section at the end of
% the paper (co-located with Acknowledgements -- the two may appear in either
% order, but both must be before References), and does not count toward the paper
% page limit. In many cases, where the ethical impacts and expected societal
% implications are those that are well established when advancing the field of
% Machine Learning, substantial discussion is not required, and a simple
% statement such as the following will suffice:

This paper presents work whose goal is to advance the field of Machine
Learning. There are many potential societal consequences of our work, none
which we feel must be specifically highlighted here.

% The above statement can be used verbatim in such cases, but we encourage
% authors to think about whether there is content which does warrant further
% discussion, as this statement will be apparent if the paper is later flagged
% for ethics review.

% % In the unusual situation where you want a paper to appear in the
% % references without citing it in the main text, use \nocite

\bibliography{HLC}
\bibliographystyle{icml2026}

% %%%%%%%%%%%%%%%%%%%%%%%%%%%%%%%%%%%%%%%%%%%%%%%%%%%%%%%%%%%%%%%%%%%%%%%%%%%%%%%
% %%%%%%%%%%%%%%%%%%%%%%%%%%%%%%%%%%%%%%%%%%%%%%%%%%%%%%%%%%%%%%%%%%%%%%%%%%%%%%%
% % APPENDIX
% %%%%%%%%%%%%%%%%%%%%%%%%%%%%%%%%%%%%%%%%%%%%%%%%%%%%%%%%%%%%%%%%%%%%%%%%%%%%%%%
% %%%%%%%%%%%%%%%%%%%%%%%%%%%%%%%%%%%%%%%%%%%%%%%%%%%%%%%%%%%%%%%%%%%%%%%%%%%%%%%
\newpage
\appendix
\onecolumn

% You can have as much text here as you want. The main body must be at most $8$
% pages long. For the final version, one more page can be added. If you want, you
% can use an appendix like this one.

% The $\mathtt{\backslash onecolumn}$ command above can be kept in place if you
% prefer a one-column appendix, or can be removed if you prefer a two-column
% appendix.  Apart from this possible change, the style (font size, spacing,
% margins, page numbering, etc.) should be kept the same as the main body.

\section{HLC with SAC - Details}\label{Appendix:HLC-SAC}
We condition our SAC Lead Critics on their respective group (\(m\) agents) observations \(o^{1:m}\) and actions \(a^{1:m}\). Following SAC, we learn two Lead Critics \(Q_{LC{\psi_{i \in \{1, 2\}}}}\) with clipped double Q-learning~\citep{ClippedDoubleQSAC} to reduce overestimation bias and target networks \(Q_{LC{\bar{\psi}_{i \in \{1, 2\}}}}\) to stabilize training~\citep{TargetNetworks}, per group of agents. For the lead critic entropy, we follow~\citep{HASAC}, where the centralized critic learns its own entropy coefficient. With a sampled minibatch \(B \sim D\) containing \((o_{t}^{1:m}, a_{t}^{1:m}, r_{t}, o_{{t+1}}^{1:m})\), we calculate the Lead Critic loss:

\begin{equation}\label{eq:LC_Critic_Update}
\begin{aligned}
      \mathcal{L}_{Q_{LC_{\psi_i}}} = 
      \mathbb{E}_{B \sim \mathcal{D},
      \text{ }a_{t+1}^{1:m} \sim \pi_{\theta^{1:m}}^{1:m}} 
      \left[ \left( Q_{LC_{\psi_i}}(o_t^{1:m}, a_t^{1:m}) - y_t \right)^2 \right] \\
     y_t = r_{LC_t} + \gamma \left(\min_{j \in \{1, 2\}} Q_{LC_{\bar{\psi_j}}}(o_{t+1}^{1:m}, a_{t+1}^{1:m}) + H_{LC}(o_{t+1}^{1:m}, a_{t+1}^{1:m}) \right) \\
    \text{with } H_{LC}(o_{t+1}^{1:m}, a_{t+1}^{1:m}) = - \alpha_{LC} 
    \cdot
    \sum_{l=1}^m{\log \pi^l_{\theta^l}(a_{t+1}^{l} | o_{t+1}^{l}) }
\end{aligned}
\end{equation}

\section{Transformer-Encoder based Lead Critic} \label{Appendix:TransformerEncoder}
\begin{figure}[ht]
  \vskip 0.2in
  \begin{center}
    \centerline{\includegraphics[width=0.7\columnwidth]{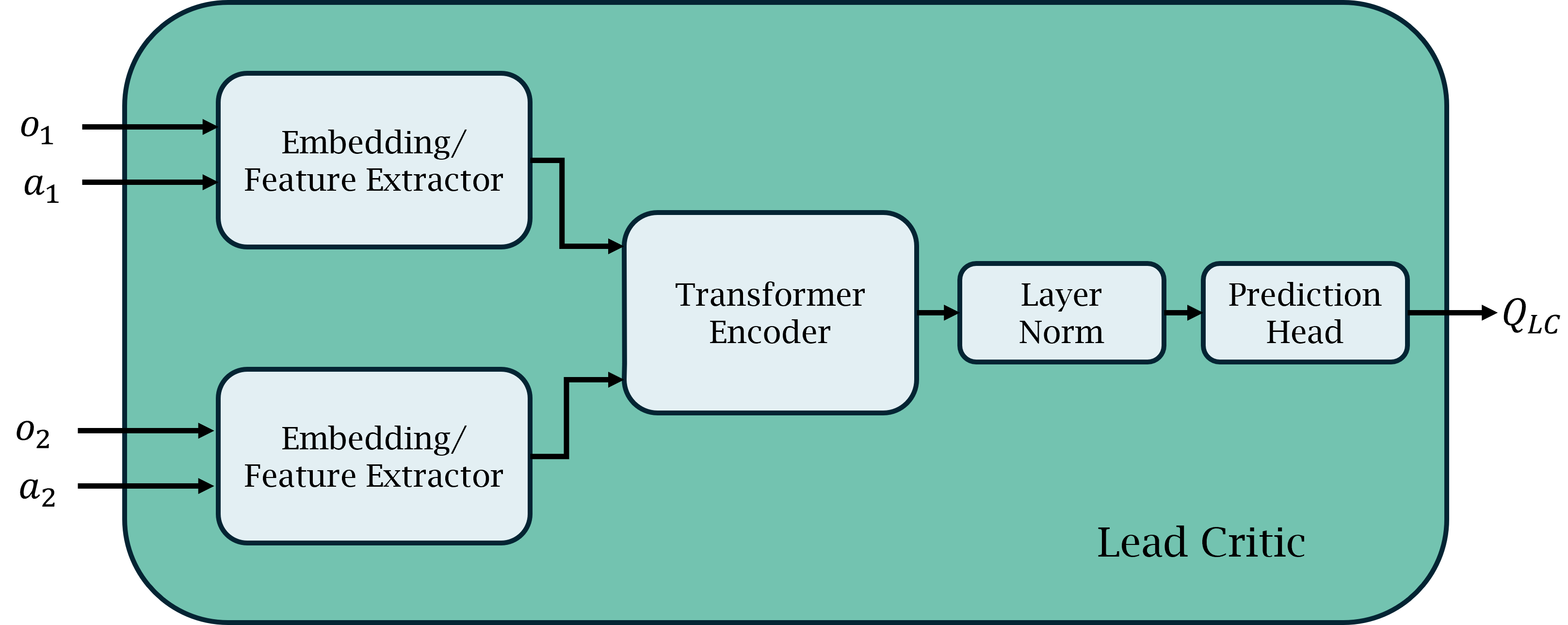}}
    \caption{
      Transformer-Encoder based Lead Critic grouping two agents
    }
    \label{fig:LC_Architecture}
  \end{center}
\end{figure}

We design the Lead Critic generally as a Transformer-Encoder based network with a 2-layer MLP prediction head with ReLU activation. Each agents' observation and action are embedded or processed by a feature extractor before being given to the Transformer-Encoder. This separate processing allows handling heterogeneous observation and action spaces. We follow~\citep{StabilizingTransformersRL} in repositioning the layer norm in the input stream of the Transformer-Encoder submodules for improved stabilization and direct gradient flow from output to input. In Figure~\ref{fig:LC_Architecture}, this Lead Critic structure is visualized for handling two agents. Hyperparameters for Transformer-Encoder based Lead Critics used in the experiments can be found in Table~\ref{tab:HLC_hp}.

\section{Sequential Updates}\label{Appendix:SequentialUpdates}
\begin{figure*}[tb]
\begin{centering}
    \subfloat[Escort8]
        {\includegraphics[scale=0.3]{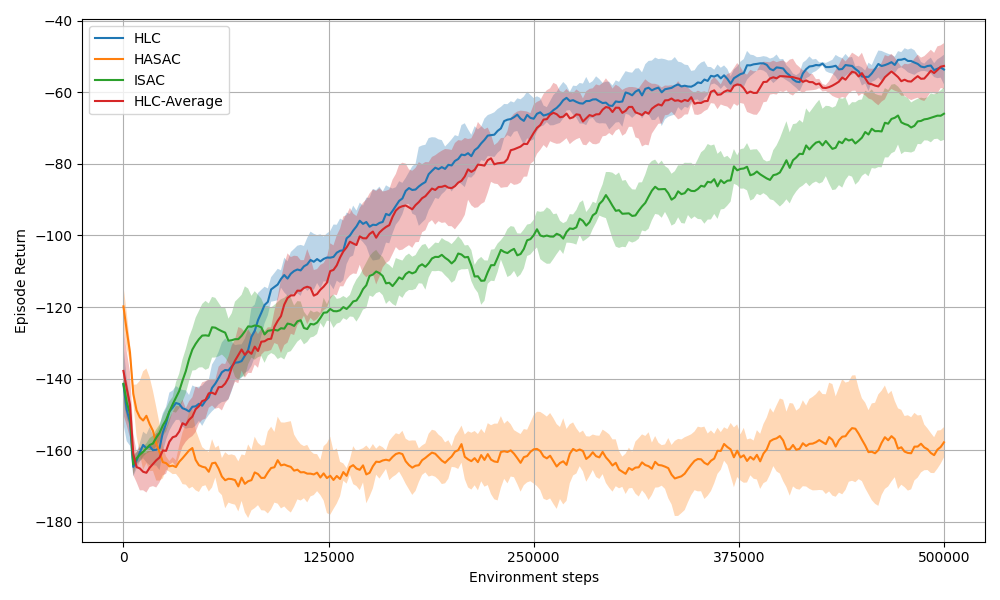}\label{subfloat:AppendixAverageEscort8}}
    \subfloat[Surveillance4]
        {\includegraphics[scale=0.3]{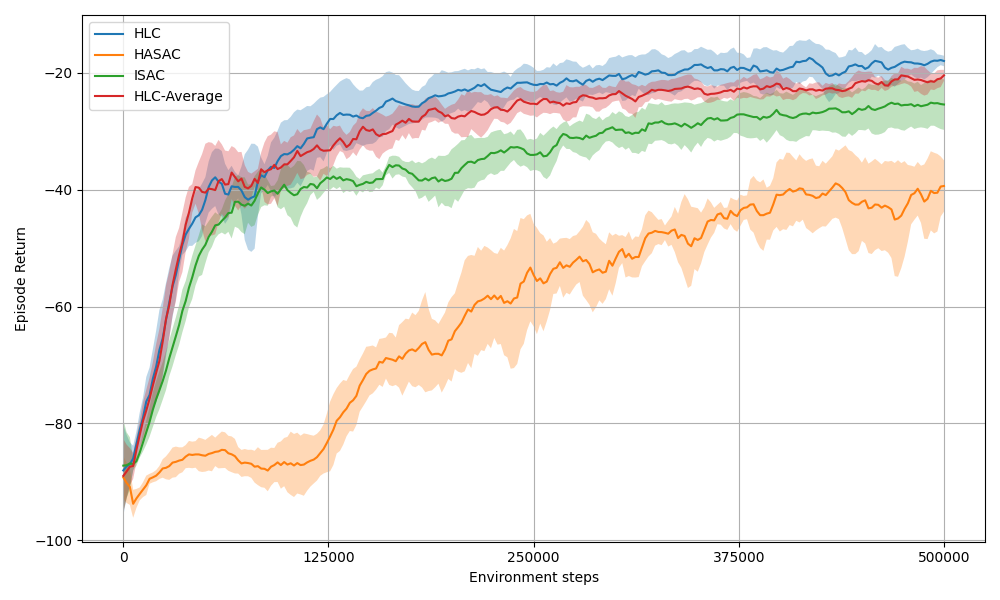}\label{subfloat:AppendixAverageSurveillance4}}
    \caption{
    HLC compared to HLC-Average
    }
    \label{fig:AppendixAverage}
\end{centering}
\end{figure*}

We provide a discussion about the differences between the HLC sequential updating scheme and a combined/averaged update strategy for updating an actor from multiple critics. In the HLC actor update, we use multiple critics to update a single actor and inference a new action after each policy update (see Figure~\ref{fig:HLC-Sequential}). Therefore, the next update can adapt to the already updated policy. In comparison, when calculating updates from multiple critics simultaneously and averaging them, the critics cannot adapt to the updates proposed by each other. This can result in conflicting gradients and HLC avoids this problem by design.\\
We compare HLC with an averaged update version termed HLC-Average, where each update is still calculated separately, but the average of gradients is used to update the actor instead of sequential updates. No other components were altered. We report results in Figure~\ref{fig:AppendixAverage}. In general, the HLC-Average variant learns slower compared to HLC and with higher variance. These findings support the argument of conflicting gradients in HLC-Average that negatively impact the training with instability and less effective updates.
Additionally, on Surveillance4 (Figure~\ref{subfloat:AppendixAverageSurveillance4}), HLC clearly outperforms HLC-Average in final performance, showcasing the benefits of the sequential updates. Notably, HLC-Average still outperforms the single hierarchy baselines. This validates the underlying HLC concept of leveraging critics from multiple hierarchies to effectively combine local optimization with global cooperation.

\section{Additional Results}\label{Appendix:AdditionalResults}
In this section, we show the results of ablations regarding different configurations of the SOTA baseline HASAC on Escort3 task, results for the Escort5 task and a comparison for HASAC and ISAC, when increasing the gradient steps on the Escort8 task. We denote the combination of X "gradient steps" with a value of Y for the "actor update interval" as GXIY. The original HASAC~\citep{HASAC} setup with centralized information for actor and critic is denoted as HASAC-Global:
\begin{itemize}
    \item HASAC-G4I2,
    \item HASAC-GlobalCritic-G1I1, centralized MLP critic with concatenated observations
    \item HASAC-GlobalCritic-G4I2, centralized MLP critic with concatenated observations
    \item HASAC-Global-G1I1, concatenated observations as actor and centralized MLP critic input
\end{itemize}

\begin{figure*}[h]
\begin{centering}
    \subfloat[Escort3 - HASAC variants]
        {\includegraphics[scale=0.22]{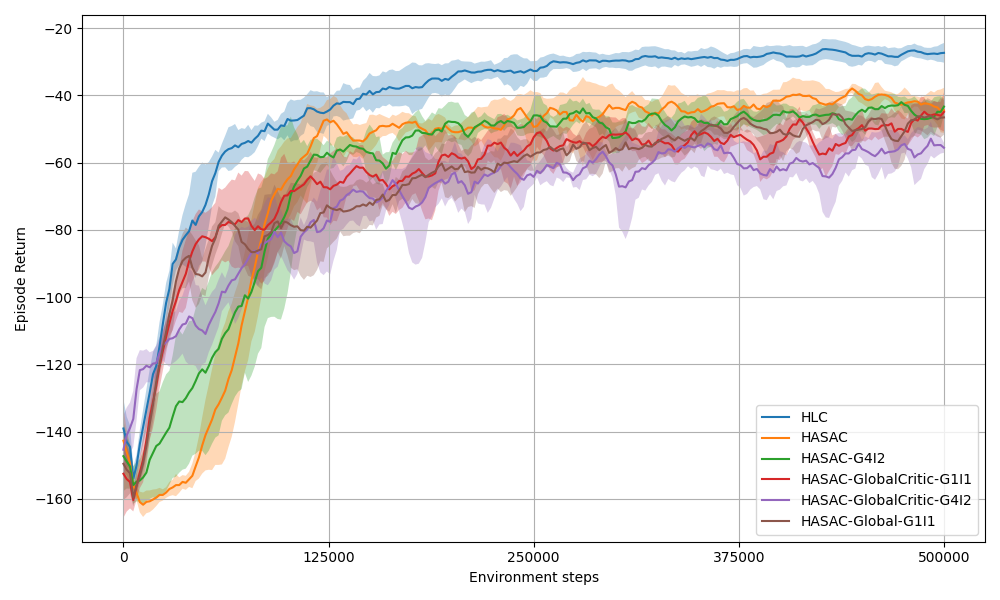}\label{subfloat:Escort3_200_HASAC_Ablations}}
    \subfloat[Escort5]
        {\includegraphics[scale=0.22]{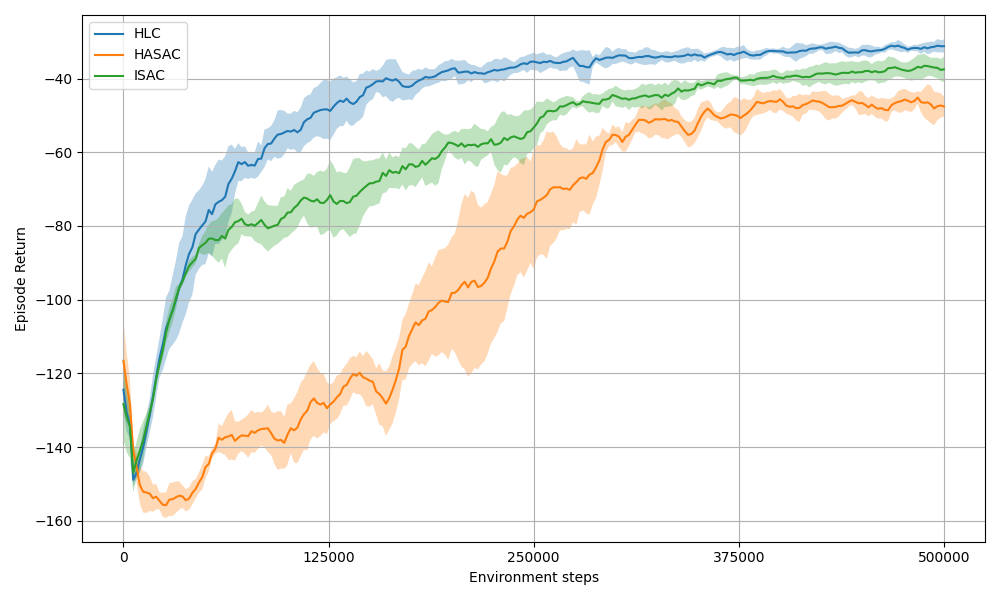}\label{subfloat:Escort5_200_Main_Result}}
    \subfloat[Escort8 - Gradient steps]
        {\includegraphics[scale=0.22]{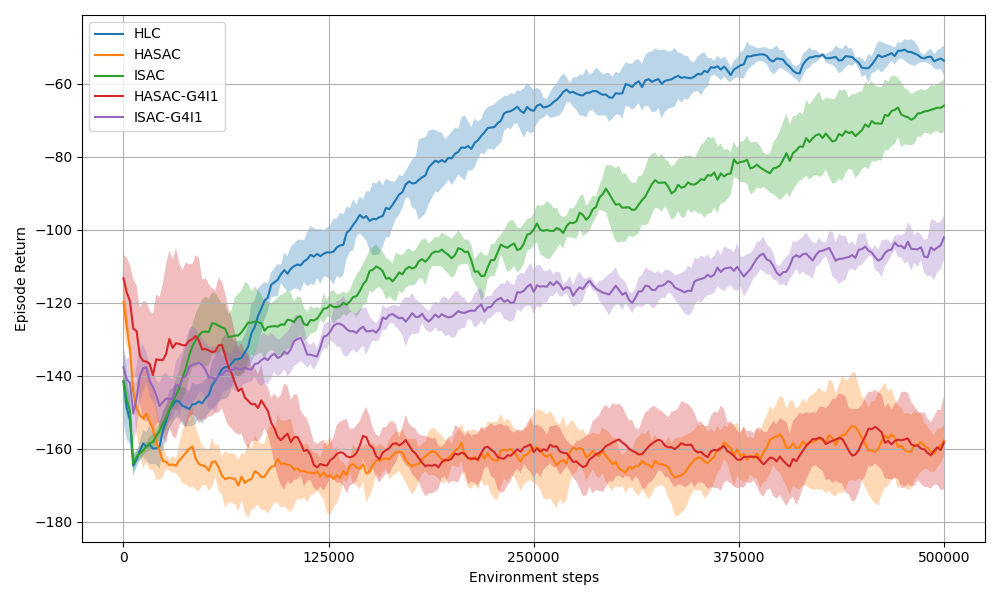}\label{subfloat:Escort8_500_G_Ablations}}\\
    \subfloat[Escort3 -HASAC variants- Episode length]
        {\includegraphics[scale=0.22]{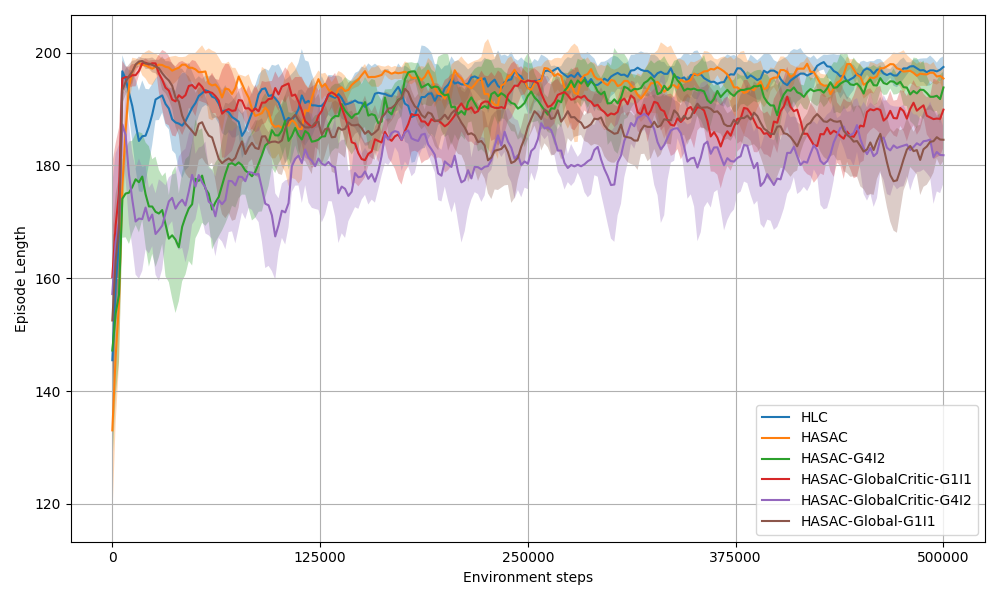}\label{subfloat:Escort3_200_length_HASAC_Ablations}}
    \subfloat[Escort5 - Episode length]
        {\includegraphics[scale=0.22]{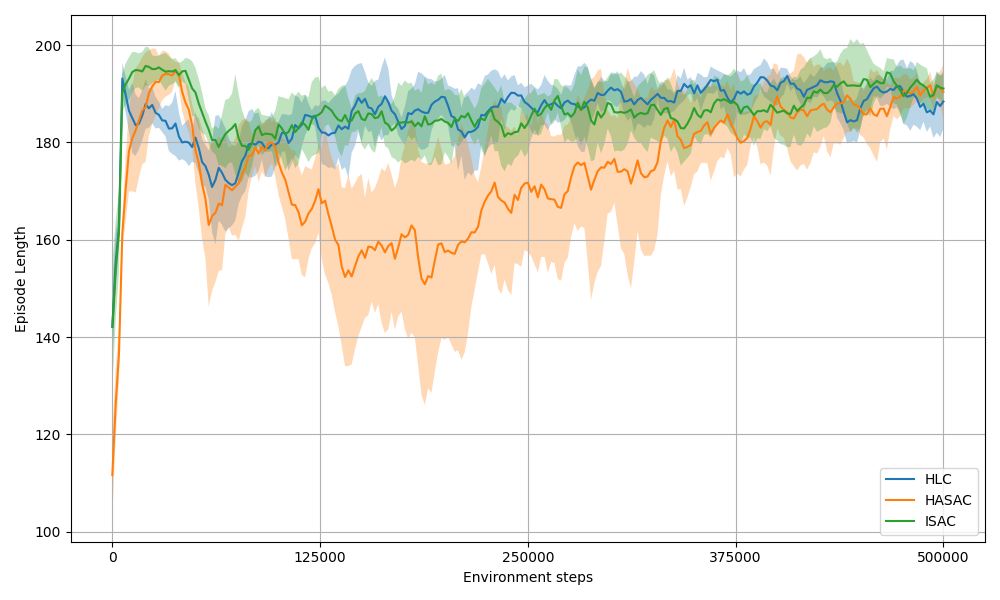}\label{subfloat:Escort5_200_length_Main_Result}}
    \subfloat[Escort8 - Gradient steps - Episode length]
        {\includegraphics[scale=0.22]{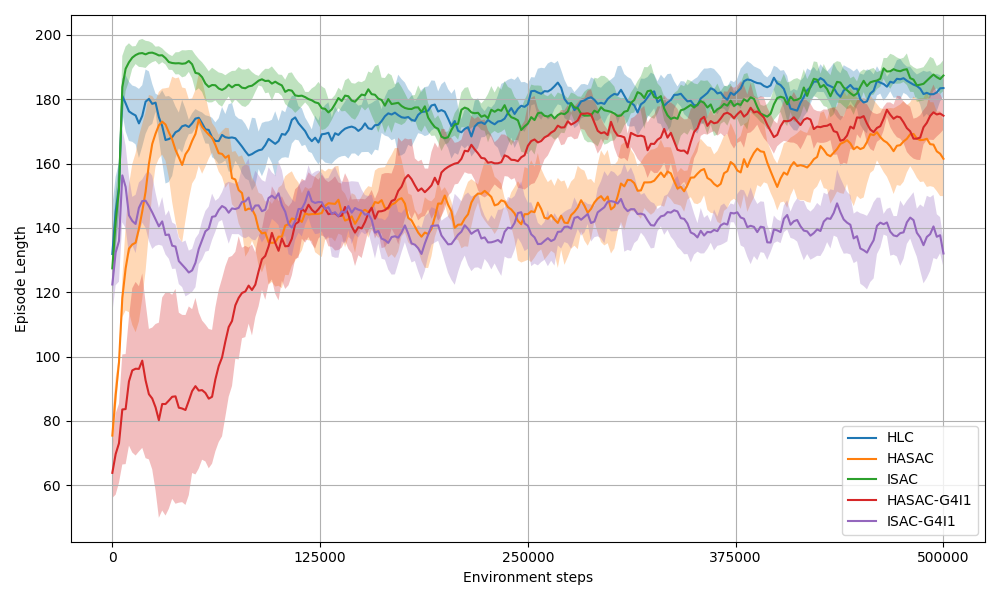}\label{fig:Escort8_500_length_G_Ablations}}
    \caption{
    Additional results for HASAC variants on Escort3, Escort5 and increasing gradient steps for HASAC and ISAC on Escort8.
    }
    \label{fig:Main_Results}
\end{centering}
\end{figure*}

\textbf{HASAC variants} As shown in Figure~\ref{subfloat:Escort3_200_HASAC_Ablations}, employing the Lead Critic architecture as proposed with HLC on HASAC results in consistently better performance compared to a global state given to a centralized MLP critic. Note that even when conditioning the HASAC actor directly on a global state, HLC still outperforms it significantly. In addition, scaling the gradient steps to the HLC setup with 4 gradient steps and an actor update interval of 2 does not reveal itself as effective, moreover it destabilizes the learning process. This showcases that HLC, with its multi perspective optimization, can effectively exploit low amounts of samples.

\textbf{Escort5} On the Escort5 task HLC also outperforms the baselines with a similar pattern as observed for Escort3 and Escort8.

\textbf{Escort8 - Baseline gradient steps} For comparison, we also ablate increasing the gradient steps for HASAC and ISAC on the Escort8 task with death penalty 500. The increase of gradient steps does not transfer to an improved performance for the baselines. Moreover it degrades the performance and slows the learning process.

\section{Experimental Details} \label{Appendix:ExperimentalDetails}
\subsection{Experimental Setup}
All experiments were conducted with 5 seeds and we report the mean performance with 95\% confidence interval.
\subsubsection{SimpleSpread}
In SimpleSpread~\cite{MPE}, agents are globally rewarded based on the sum of minimum distances from all agents to landmarks and locally penalized for collisions with other agents. The final reward is an average of global and local rewards. Agents observe their position, velocity, relative position, and velocities of other agents and landmarks. For constructing a cooperative reward, we follow~\citep{HASAC} by summing the rewards of all agents. We give this cooperative reward to centralized critics in HLC and HASAC, while local critics from HLC and ISAC receive the default reward. Each algorithm is trained for 2 million environment steps and evaluated every 20000 environment steps.

\subsubsection{Momaland Escort and Surveillance}
In the proposed Escort and Surveillance tasks, each algorithm is trained for 500 thousand environment steps and evaluated every 2000 environment steps. Reward calculation can be found in \ref{sec:MomalandTasks}.\\
Here, we also provide the list of changes to MOMAland~\cite{felten2024momaland} version 0.1.1. The code is provided in the supplementary material and the folder containing the extended environments are denoted as "escort\_custom" and "surveillance\_custom", with the modified base environment file as crazyRL\_base\_custom.py".
 We introduced:
\begin{itemize}
    \item Seeding
    \item Agents and target positions randomized on resets
    \item Target movement with multiple random direction changes per episode
    \item Increased map size
    \item Relative position observations of other agents and the target
    \item Only ego height observation, no xy information
    \item Normalized reward components and scaled termination penalty
\end{itemize}
For all tasks we use a map size of 8 and 4 random target direction changes within an episode.\\
We normalize our rewards with the map size and clip them between [-1,0] as described in~\autoref{eq:reward_normalization} with \(r^i_x\) representing any reward component.

\begin{equation}\label{eq:reward_normalization}
\begin{aligned}
      r^i_x = clip(\frac{r^i_x}{map\text{ }size)}, -1, 0)
\end{aligned}
\end{equation}

\subsection{Training Details and Hyperparameters}

\begin{table}[ht]
    \centering
    \caption{Common hyperparameters in all environments for all algorithms}
    \begin{tabular}{@{}l c}
        \toprule
        \textbf{Hyperparameter} & \textbf{Value} \\ \midrule
        Parallel environments \(P\)      & 10        \\
        Replay buffer size          & 1e6       \\
        Discount factor             & 0.99      \\
        Polyak coefficient          & 0.005     \\ 
        Adam \(\epsilon\)           & 1e-5      \\
        Batch size                  & \(256 \cdot P\)    \\
        \bottomrule
    \end{tabular}
    \label{tab:all_envs_hp_common}
\end{table}

\begin{table}[ht]
    \centering
    \caption{Environment specific hyperparameters for all algorithms. With EscortX for all Escort tasks.}
    \begin{tabular}{@{}l c c}
        \toprule
        \textbf{Hyperparameter} & \textbf{EscortX} & \textbf{SimpleSpread} \\ \midrule
            Train interval & 20 & 50    \\
            Total env steps & 5e5 & 2e6 \\
            Warmup steps & 5e3 & 1e4 \\
            Observation history & 2 & 1 \\
            n-step & 5 & 2 \\
            Actor lr & 3e-4 & 5e-4 \\
            Critic lr & 1e-3 & 5e-4 \\
            Lead Critic lr & 1e-3& 5e-4 \\
            Auto alpha & True & False \\
            Alpha lr & 3e-4 & - \\
            Alpha init   & 1.0 & 0.2 \\
            \bottomrule
    \end{tabular}
    \label{tab:specific_envs_hp_common}
\end{table}

\begin{table}[ht]
    \centering
    \caption{HLC specific hyperparameters in all environments}
    \begin{tabular}{@{}l c}
        \toprule
        \textbf{Hyperparameter} & \textbf{Value} \\ 
        \midrule
        Actor network type                  & HLC Actor \\
        Actor optimizer                     & RAdam~\cite{RAdam}     \\
        Actor optimizer weight decay        & 1e-5      \\
        Subnetworks                         & 2         \\
        Subnetwork hidden sizes             & [256, 256]\\
        Subnetwork activation               & ReLU      \\
        Base network hidden sizes              & [256, 256]\\
        Base network activation                & ReLU      \\
        Actor attention embedding dim       & 128        \\
        Actor attention heads               & 4         \\
        Critic network type                 & MLP     \\
        Critic hidden sizes        & [256, 256]    \\
        Critic activations         & ReLU              \\
        Critic optimizer                    & Adam~\cite{Adam}     \\
        Lead critic network type            & Transformer-Encoder \\
        Lead critic optimizer               & RAdam~\cite{RAdam}     \\
        Lead critic optimizer weight decay  & 1e-5      \\
        Lead critic embedding dim           & 128        \\
        Lead critic number Transformer blocks  & 2         \\
        Lead critic attention heads         & 4         \\
        Lead critic Transformer feedforward dim & 512   \\
        Gradient steps              & 4         \\
        Gradient steps (SimpleSpread) & 2\\
        Actor update interval       & 2         \\
        \bottomrule
    \end{tabular}
    \label{tab:HLC_hp}
\end{table}

\begin{table}[ht]
    \centering
    \caption{Common hyperparameters for HASAC}
    \begin{tabular}{@{}l c}
        \toprule
        \textbf{Hyperparameter} & \textbf{Value} \\ \midrule
        Actor network type          & MLP \\
        Actor optimizer             & Adam~\cite{Adam}        \\ 
        Actor hidden sizes          & [256, 256]    \\
        Lead critic network type            & Transformer-Encoder \\
        Lead critic optimizer               & RAdam~\cite{RAdam}     \\
        Lead critic optimizer weight decay  & 1e-5      \\
        Lead critic embedding dim           & 128        \\
        Lead critic number Transformer blocks  & 2         \\
        Lead critic attention heads         & 4         \\
        Lead critic Transformer feedforward dim & 256   \\
        Gradient steps              & 1         \\
        Actor update interval       & 1         \\
        \bottomrule
    \end{tabular}
    \label{tab:HASAC_hp}
\end{table}

\begin{table}[ht]
    \centering
    \caption{Common hyperparameters for ISAC}
    \begin{tabular}{@{}l c}
        \toprule
        \textbf{Hyperparameter} & \textbf{Value} \\ \midrule
        Actor network type          & MLP \\
        Actor optimizer             & Adam~\cite{Adam}        \\ 
        Actor hidden sizes          & [256, 256]    \\
        Critic network type                 & MLP     \\
        Critic hidden sizes        & [256, 256]    \\
        Critic activations         & ReLU              \\
        Critic optimizer                    & Adam~\cite{Adam}     \\
        Gradient steps              & 1         \\
        Actor update interval       & 1         \\
        \bottomrule
    \end{tabular}
    \label{tab:ISAC_hp}
\end{table}

We report the common hyperparameters for all algorithms in Table~\ref{tab:all_envs_hp_common} and Table~\ref{tab:specific_envs_hp_common}. Specific hyperparameters for our approach HLC are reported in Table~\ref{tab:HLC_hp}. For HASAC, we use a 2-layer MLP actor with hidden dimension 256 and ReLU activations and for variants with a global MLP critic, we also employ a 2-layer MLP network. For HASAC ablations, the changes to this setup are described at the respective experiments. The basic HASAC hyperparameters are given in Table~\ref{tab:HASAC_hp}.
In ISAC, we also use a 2-layer MLP for actor and critic with hidden dimension 256 and ReLU activations. Both actor and critic only receive the agents' local observations (and actions) as input. ISAC hyperparameters are given in Table~\ref{tab:ISAC_hp}.\\
In general, the chosen hyperparameters are based on common values found in the literature and we have purposely avoided tuning HLC to specific environment conditions.
\section{Video results}\label{Appendix:VideoResults}
\begin{figure}[ht]
  \vskip 0.2in
  \begin{center}
    \centerline{\includegraphics[width=0.8\columnwidth]{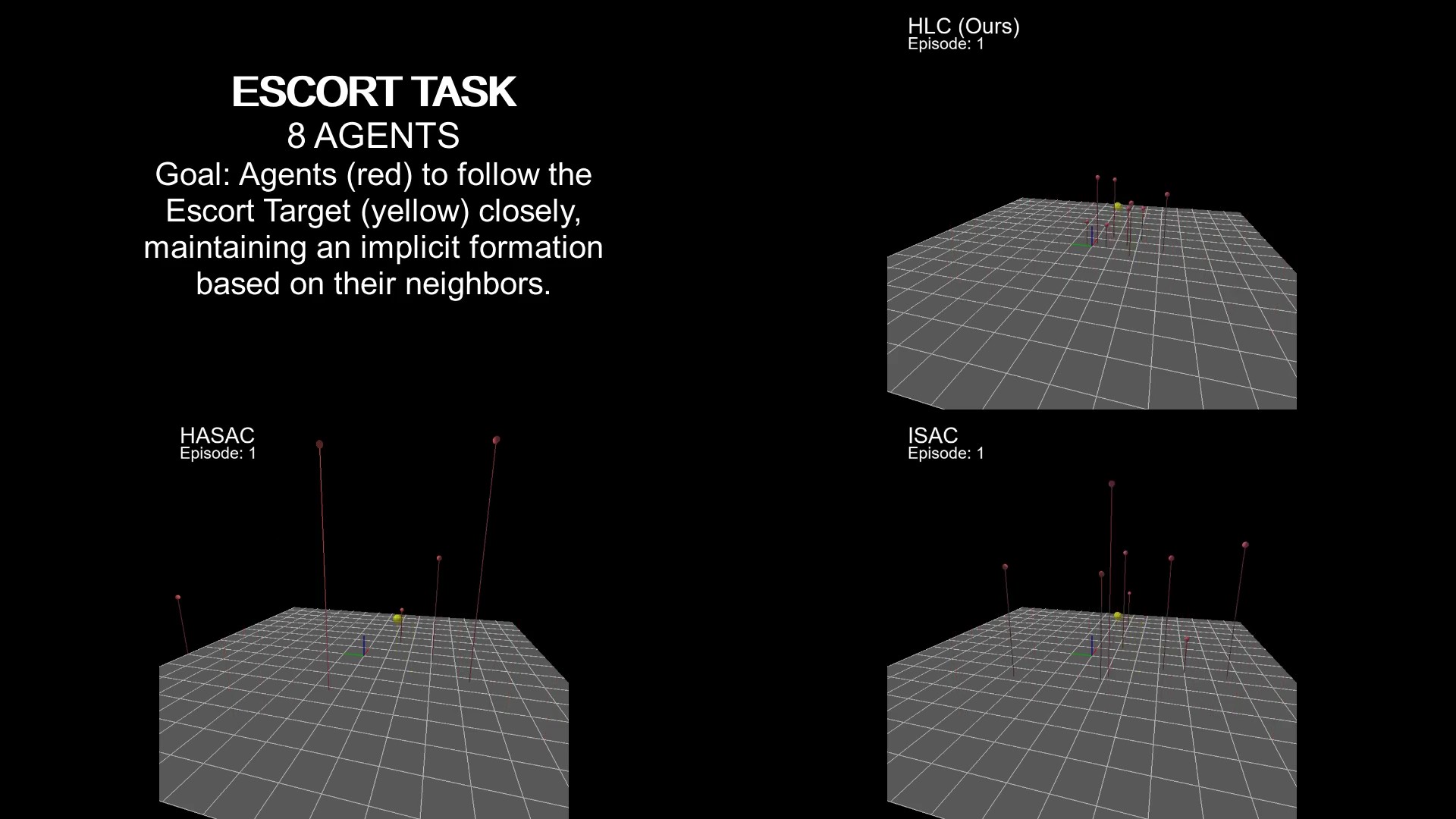}}
    \caption{
      Single frame of video provided in the supplementary material
    }
    \label{fig:HLC_Videos_single_frame}
  \end{center}
\end{figure}

We provide videos of agents on the Escort3, Escort8 (death penalty 500) and Surveillance4 tasks within the supplementary material for HLC, HASAC and ISAC. Figure~\ref{fig:HLC_Videos_single_frame} shows an exemplary frame of the provided video. Note that in the video we show only 5 episodes for demonstration purposes.
%%%%%%%%%%%%%%%%%%%%%%%%%%%%%%%%%%%%%%%%%%%%%%%%%%%%%%%%%%%%%%%%%%%%%%%%%%%%%%%
%%%%%%%%%%%%%%%%%%%%%%%%%%%%%%%%%%%%%%%%%%%%%%%%%%%%%%%%%%%%%%%%%%%%%%%%%%%%%%%

\end{document}